\documentclass[10pt,twocolumn,letterpaper]{article}

\usepackage[pagenumbers]{wacv} % To force page numbers, e.g. for an arXiv version
\usepackage{colortbl}
\usepackage{xcolor}
\usepackage{lineno}
\definecolor{fail}{gray}{0.9}
\usepackage{graphicx}
\usepackage{booktabs}
\usepackage{multirow}
\definecolor{fail}{gray}{0.9}

\usepackage{tikz}
\usetikzlibrary{arrows.meta, positioning, shapes.geometric, fit}

\usepackage[accsupp]{axessibility}  % Improves PDF readability for those with disabilities.

\usepackage{algorithm}
\usepackage{algorithmic}
\usepackage{lineno}
\usepackage{textcomp}
\usepackage{booktabs}
\usepackage{float}
\usepackage{silence}

\definecolor{wacvblue}{rgb}{0.21,0.49,0.74}
\usepackage[pagebackref,breaklinks,colorlinks,allcolors=wacvblue]{hyperref}

\def\wacvPaperID{3758} % *** Enter the WACV Paper ID here
\def\confName{WACV}
\def\confYear{2027}

\begin{document}

%%%%%%%%% TITLE - PLEASE UPDATE
\title{Spike-Aware Gradient-Based Mixed-Precision Quantization of Spiking Neural Networks for Vision Applications}

\author{
Rajshekhar Rakshit\\
Indian Institute Of Technology Madras\\
Chennai, India\\
{\tt\small cs24s031@cse.iitm.ac.in}
\and
Devgokul B.V\\
Indian Institute Of Technology Madras\\
Chennai, India\\
{\tt\small ns26z056@smail.iitm.ac.in}
\and
Rakshitha Kalkura\\
Indian Institute Of Technology Madras\\
Chennai, India\\
{\tt\small cs25e052@smail.iitm.ac.in}
\and
Gopalakrishnan Srinivasan\\
Indian Institute Of Technology Madras\\
Chennai, India\\
{\tt\small sgopal@cse.iitm.ac.in}
}
\maketitle

%%%%%%%%% ABSTRACT
\begin{abstract}
   In recent years, Spiking Neural Networks (SNNs) have gained massive attention as an energy-efficient neural computing paradigm due to their event-driven and sparse computing capability. However, maintaining accuracy under low-precision constraints, necessary for practical deployment, has been largely unexplored for SNN models. In addition, existing Hessian-based approaches for quantization, such as HAWQ-V2, incur significant memory and computational overhead due to temporal unrolling across multiple SNN timesteps. We propose a spike-aware gradient-based mixed-precision quantization framework for SNNs that adaptively assigns bit-widths using first-order gradient information, avoiding costly second-order Hessian computations. We evaluate our method across various SNN-based architectures, including VGG, ResNet, and Vision Transformer models obtained from both ANN-to-SNN conversion and direct training frameworks, on CIFAR10/100, ImageNet, and DVS-CIFAR-10 datasets. Notably, on ResNet-34 evaluated on ImageNet, our method achieves 73.30\% accuracy at 4.78$\times$ compression, which is comparable to that provided by HAWQ-V2, while reducing memory consumption by 2.31$\times$ and bit-allocation runtime by 100$\times$. We also present an energy analysis for mixed-precision SNNs showing that the proposed quantization strategy enables energy-efficient deployment without compromising performance.
\end{abstract}

%%%%%%%%% BODY TEXT
\section{Introduction}
\label{sec:intro}

Spiking Neural Networks or SNNs~\cite{gerstner2002spiking}, which are motivated by the computational efficiency of the human brain, have emerged as an energy-efficient alternative to state-of-the-art Artificial Neural Networks (ANNs). Unlike ANNs, which perform dense multiply-accumulate (or MAC) operations on continuous-valued activations, SNNs process information using discrete and binary spikes that a neuron emits only when its membrane potential crosses a firing threshold. Since computations are triggered by spikes, this event-driven behavior makes neural activity sparser in both space and time, allowing the MAC operations in ANNs to be replaced by much cheaper accumulate (AC) operations. Neuromorphic hardware architectures~\cite{loihi, truenorth, kadway2023akida, painkras2013spinnaker} are designed to take advantage of this sparse, event-driven, and spike-based computation, enabling SNNs to realize their energy-efficiency benefits during deployment.

Recent advances in SNN training techniques~\cite{ramesh2025pascal, bu2022optimal, deng2022tet, 8891809} have enabled them to achieve inference-time accuracy competitive with their ANN counterparts on a range of vision datasets, making them increasingly viable candidates for real-world deployment. However, realizing the practical benefits of SNNs still requires reducing their memory footprints and computational budgets. Methods such as quantization are therefore necessary to realize efficient deployment of SNNs on resource-constrained edge and neuromorphic hardware.
% \vspace{-0.3em}
% \begin{figure}[!ht]
%     \centering
%     \includegraphics[width=\columnwidth]{plots/memory_resnet18.png}
%     \caption{Peak GPU memory needed for ANN v/s SNN by HAWQ-V2 and Our method for ResNet18 on CIFAR10 dataset }
%     \label{fig:peak_gpu_mem}
% \end{figure}
\begin{figure}[!t]
    \centering
    \includegraphics[width=\columnwidth]{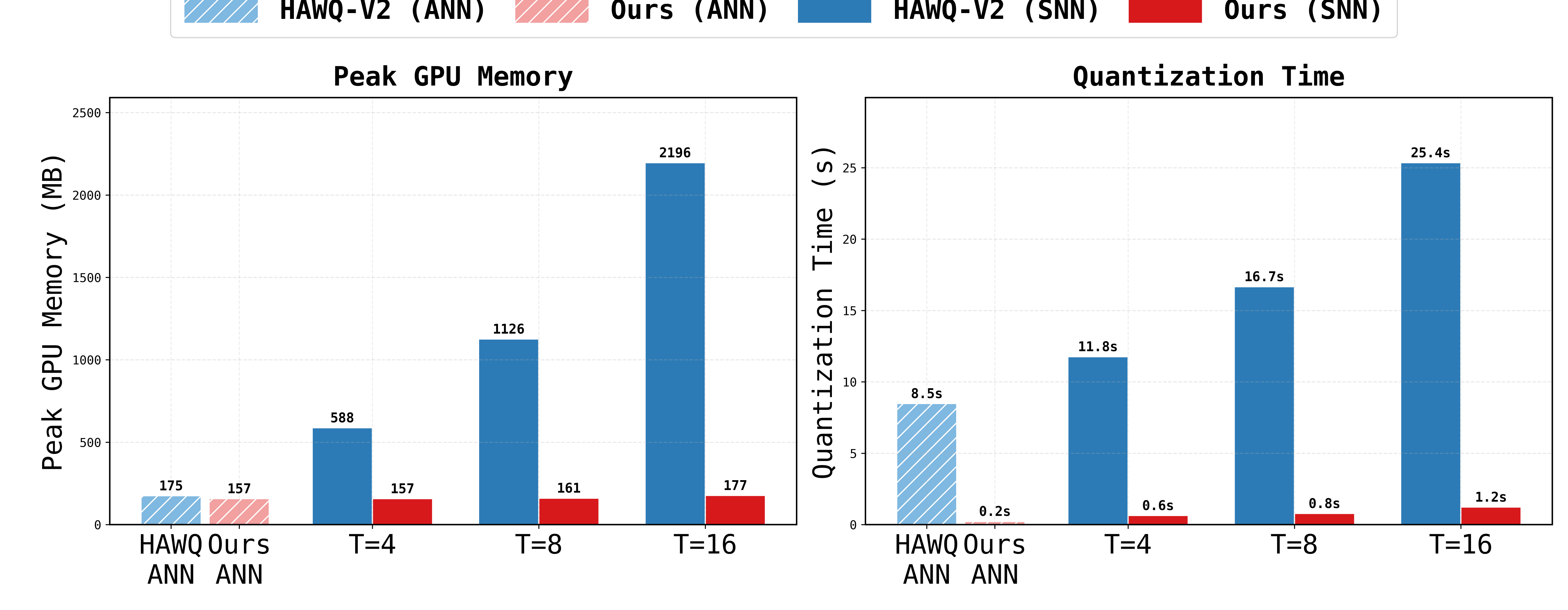}
    \caption{Peak GPU memory consumption and quantization time of ResNet-18, on CIFAR-10, using HAWQ-V2 and the proposed method for both ANN and SNN variants across different timestep configurations, measured on an NVIDIA RTX 4090 GPU.}
    \label{fig:comparison_plot}
\end{figure}

Quantization is an effective technique to reduce the model size and computational cost by using fewer bits to represent parameters and activations~\cite{Jacob_2018_CVPR, han2016deep}. Existing SNN quantization methods primarily use uniform bit-widths across layers~\cite{10.1145/3664647.3681186,wei2025qpsnn}. However, it is now well established that the layers in a neural network differ substantially in their sensitivity to quantization noise~\cite{Dong_2019_ICCV}. This motivates mixed-precision quantization, where bit-widths are assigned according to individual layer sensitivity.

Among existing methods, HAWQ-V2 \cite{NEURIPS2020_d77c7035} estimates such sensitivity using second-order Hessian information with the Hutchinson trace estimator. However, applying this directly to SNNs is computationally expensive because BPTT~\cite{Werbos1990BackpropagationTT} unrolls the network over time, requiring Hessian-vector products over the temporally expanded graph and increasing both computation and memory with the number of timesteps. Figure~\ref{fig:comparison_plot} compares this cost with our method for ResNet-18 on CIFAR-10, showing rapid increase in HAWQ-V2's runtime and memory footprint as the number of timesteps grows. % In contrast, the proposed method requires only a single first-order backward pass, resulting in nearly constant resource usage across timesteps.

\textbf{Contributions.}
In this work, we propose a spike-aware gradient-based mixed-precision quantization framework for SNNs using first-order gradient information together with spike-driven activity statistics. Our key contributions are as follows.
\begin{enumerate}
    \item \textbf{A spike-aware gradient-based mixed-precision quantization framework.} The proposed method adaptively allocates bit-widths using the first-order gradient information and spike activity statistics.
    \item \textbf{An efficient first-order sensitivity estimation strategy.} The proposed method eliminates the expensive second-order Hessian computations required by HAWQ-based methods~\cite{NEURIPS2020_d77c7035, Dong_2019_ICCV}, substantially reducing the computational overhead of bit allocation while maintaining competitive quantization performance (see Figure~\ref{fig:comparison_plot}).
    \item \textbf{Comprehensive benchmarking.} We validate the proposed technique on SNN architectures such as VGG-16, ResNet-18/20/34, and vision transformer, obtained using ANN-SNN conversion and direct backpropagation-based training, on static vision (CIFAR-10, CIFAR-100, ImageNet) and neuromorphic datasets.
\end{enumerate}
% Experiments demonstrate that the proposed method consistently achieves a comparable accuracy to Hessian methods, but with much lesser runtime and memory. 
\vspace{-0.2em}
\section{Related Work and Background}
\vspace{-0.2em}
\paragraph{Spiking Neural Networks.} SNNs process information over discrete timesteps using spiking neurons that integrate weighted inputs and produce spikes when the membrane potential exceeds a threshold. They are primarily trained using either ANN-SNN conversion \cite{bu2022optimal, ramesh2025pascal} or direct training \cite{deng2022tet, 8891809}. ANN-SNN conversion first trains an ANN using standard backpropagation and then replaces activation functions with spiking neurons, where ANN activations are approximated using spike rates accumulated over multiple timesteps. Methods such as QCFS \cite{bu2022optimal} and PASCAL~\cite{ramesh2025pascal} reduce conversion error by introducing quantization-aware activations during ANN training. On the other hand, direct training optimizes SNNs directly in the spiking domain using surrogate gradients \cite{8891809} and Backpropagation Through Time (BPTT) \cite{Werbos1990BackpropagationTT}, which unrolls the network over all simulation timesteps. Recent methods such as Temporal Efficient Training (TET) \cite{deng2022tet} further improve optimization by applying supervision at every timestep. Additional details of both training methods are provided in Sec. 1 of the supplementary material.
\vspace{-0.3cm}
\paragraph{Quantization of SNNs.} 
Recent works on SNN quantization have made notable progress in reducing both the memory and computational costs. Q-SNN~\cite{10.1145/3664647.3681186} binarizes weights during training using a weight-spike dual regulation objective, while QP-SNN~\cite{wei2025qpsnn} combines uniform fixed-bit post-training quantization with structured pruning. A recent fixed heterogeneous-precision strategy~\cite{10.3389/fnins.2025.1665778} employs a W1/W2 configuration, assigning W1-bit weights to the first layer and W2-bit weights to the remaining layers. While this exploits the different data representations across layers, the precision assignment is fixed by layer position rather than quantization sensitivity.

Mixed-precision quantization instead allocates higher precision to sensitive layers and lower precision to robust ones. Prior works apply the HAWQ framework~\cite{Dong_2019_ICCV,NEURIPS2020_d77c7035, 10.1145/3477145.3477158} to SNNs using Hessian-based sensitivity estimation. However, they require expensive second-order computations and manual bit allocation, while not explicitly accounting for spike-driven temporal dynamics. In contrast, our method uses a spike-aware first-order sensitivity metric to automatically determine the layer-wise precision, avoiding Hessian computation while incorporating spiking activity into the bit-allocation process.
Hardware-oriented approaches such as QUANTISENC~\cite{10658724} support configurable layer-wise precision for efficient neuromorphic hardware, but primarily focus on hardware realization rather than automatic sensitivity-driven bit allocation.
\section{Methodology}
\vspace{-0.1em}
\subsection{Quantization Formulation}
Let $W=\{W_1,W_2,\dots,W_L\}$ denote the pretrained SNN weights, where $L$ denotes the total number of SNN layers. Given a target bit-width $b_l$ for layer $l$, uniform quantization maps floating-point weights to discrete integer levels. 
For a weight tensor $W_l$, the quantized weights are computed as
\[
W_{q,l} = s_l \cdot \text{clip}
\left(
\left\lfloor
\frac{W_l}{s_l}
\right\rceil,
-q_{\max},
q_{\max}
\right),
\]
where $s_l$ denotes the quantization scale factor and $q_{\max}=2^{b_l-1}-1$. The quantization process introduces a perturbation $\Delta W_l = W_{q,l} - W_l$, which injects quantization noise during inference.% We propose a mixed-precision quantization framework that adaptively allocates precision based on layer-wise quantization sensitivity to achieve maximal compression with minimal degradation in accuracy.

\subsection{Quantization Sensitivity Analysis}

Let $\Delta W$ be the weight perturbation introduced as a result of quantization. Based on a second-order approximation of the loss landscape around a convergent solution, the resulting increase in loss can be estimated as 
\[
\Delta L \approx \frac{1}{2}\Delta W^T H\Delta W,
\]
where $H = \nabla^2 L(W)$ is the Hessian matrix. HAWQ-V2~\cite{NEURIPS2020_d77c7035} uses Hessian-based curvature to measure layer-wise quantization sensitivity, but incurs additional overhead in temporally unrolled SNNs due to iterative trace estimation.

We propose a first-order gradient metric derived from the Fisher information matrix~\cite{JMLR:v21:17-678} to obtain a computationally efficient approximation of layer-wise curvature sensitivity. For a model trained using negative log-likelihood,
\[
L(w) = -\log p(y|x;w),
\]
the Fisher information matrix~\cite{3019383} is defined as 
\[
F =
\mathbb E
\left[
\nabla_{w}\log p(y | x;w) \nabla_{w}\log p(y | x;w)^T
\right].
\]
Under standard regularity conditions for a correctly specified likelihood model, the expected Fisher information coincides with expected Hessian of negative log-likelihood~\cite{JMLR:v21:17-678, 3019383}. However, empirical Fisher need not coincide with exact Hessian in practical optimizations \cite{Kunstner2019LimitationsOT}, particularly in surrogate-gradient-trained SNNs. We therefore use the empirical Fisher as a computationally efficient proxy for curvature sensitivity rather than assuming an exact Hessian reconstruction, which is specified as
% this equivalence does not necessarily hold numerically for surrogate-gradient-trained SNNs, where the non-differentiable spike function is replaced by a surrogate derivative during optimization. Therefore, rather than requiring an explicit Hessain computation, we use the Fisher information as a computationally efficient proxy for curvature-based sensitivity.
\[
F_{ii} \approx \mathbb{E}\left[g_i^2\right],
\text{where }
g_i = \frac{\partial L}{\partial w_i}.
\]
The expectation is estimated over calibration samples. Using the diagonal approximation, the quantization-induced loss increase becomes
\[
\Delta L = \frac{1}{2}\sum_i H_{ii}(\Delta w_i)^2.
\]
\textbf{Fisher-Hessian relation in SNNs.}
Since Fisher and Hessian are not expected to coincide exactly under surrogate-gradient training, we empirically decompose their relationship within each layer as 
\[
H_{ii} = \alpha_l F_{ii} + R_i,
\]
where $l$ denotes the layer index, and $i$ denotes the parameter index within layer $l$. Here, $H_{ii}$ and $F_{ii}$ are the diagonal Hessian and Fisher entries corresponding to parameter $i$, $R_i$ denotes their discrepancy, and $\alpha_l$ is a layer-wise scaling factor. This formulation does not assume $H=F$, instead, it allows Fisher and Hessian to differ both in scale and in their residual curvature structure. 
Substituting this into the second-order loss gives
\[
\Delta L \approx \frac{1}{2}\sum_i(\alpha_l F_{ii} + R_i)(\Delta w_i)^2.
\]
Decomposing the above equation further gives
\[
\Delta L \approx \frac{\alpha_l}{2}\sum_iF_{ii}(\Delta w_i)^2 + \frac{1}{2}\sum_i R_i(\Delta w_i)^2.
\]
The first-term is the Fisher-captured sensitivity metric and the second-term is the residual curvature. Importantly, this decomposition is used only to analyze the relation between Fisher and Hessian, and neither $H$, $\alpha_l$, nor $R$ is required by the proposed quantization procedure. In the proposed method, the first term is used as a computationally efficient sensitivity proxy, which can be obtained entirely from first-order gradients. The residual term therefore characterizes the curvature component not captured by the Fisher approximation rather than introducing an additional computational requirement. Additional details are provided in Sec. 5 of the supplementary material.

\subsection{Gradient-Based Mixed-Precision Metric}
\label{sec:gbmp}

Based on the above formulation, we define the layer-wise quantization sensitivity metric as the Fisher-weighted quantization perturbation energy, which is given by
\[
\Omega_l = \sum_i g_{l,i}^2 (\Delta w_{l,i})^2
\]
where $g_{l,i}$ denotes the gradient corresponding to the $i-th$ parameter of layer $l$, and $\Delta w_{l,i}=w_{q,l,i}-w_{l,i}$ represents the quantization perturbation introduced by low-bit quantization. The proposed formulation jointly captures parameter sensitivity through squared gradients and quantization distortion through perturbation magnitude. To avoid bias towards layers with large gradient magnitudes, the sensitivity score is normalized as
\vspace{-0.1em}
\[
S_l = \frac{\sum_i  g_{l,i}^2 (\Delta w_{l,i})^2}{\sum_i g_{l,i}^2 + \epsilon}.
\]
The normalization term eliminates scale dependence across layers and enables a fair comparison of quantization sensitivity.

\subsection{Threshold Margin-Aware Bit Allocation}
\label{sec:threshold_margin_aware_alloc}
Although our gradient-based sensitivity metric captures the effect of quantization on the optimization landscape, it does not explicitly account for the temporal dynamics unique to SNNs. To incorporate the spiking neuronal dynamics into the bit allocation procedure, we estimate a layerwise membrane threshold margin during the calibration forward pass.
For every neuron, the membrane threshold margin is defined as the normalized distance between its membrane potential and the firing threshold, as specified by
\vspace{-0.2em}
\[
d_{j,t} = \frac{|V_{th} - V_{j,t}|}{V_{th}},
\]
where $V_{j,t}$ is the membrane potential and $V_{th}$ is the neuronal firing threshold.
The threshold margin is accumulated over all neurons and simulation timesteps to obtain a layer-wise estimate, formulated as
\vspace{-0.2em}
\[
m_l = \frac{1}{NT}\sum_{t=1}^{T}\sum_{j=1}^N d_{j,t}.
\]
The resulting layer-wise margins are normalized to the interval [0,1], enabling consistent comparison across layers. 
A smaller threshold margin indicates that the membrane potential is closer to the neuronal firing threshold; such neurons are more susceptible to quantization-induced perturbations since small weight changes could alter spike generation. These layers are treated more conservatively during bit allocation.

\subsection{Mixed-Precision Bit Allocation}
Given the layer-wise sensitivity scores computed as detailed in \cref{sec:gbmp}, we formulate the mixed-precision allocation as a constrained optimization problem: minimize the total sensitivity cost across all layers subject to a global bit budget. Formally, let $b_{l}$ denote the bit-width assigned to layer $l$, $n_l$ be the number of quantizable parameters in layer $l$, $B$ the total bit budget, and $C$ the set of candidate bit-widths. The optimization objective can be formally described by 
\vspace{-0.2em}
\[\min_{\{b_l\}}\sum_{l=1}^{L}S_l(b_l) \text{ subject to } \sum_{l=1}^{L} n_l . b_l \le B, \quad b_l \in \mathcal{C}, \]
where $S_l(b_l)$ is the sensitivity score of layer $l$ quantized to $b_l$ bits. Layers with high sensitivity scores are assigned larger bit-widths, while robust layers are quantized more aggressively to reduce memory. The budget $B$ is a single global constraint; the network as a whole must not exceed the prescribed bit-budget, but individual layers are free to use different precisions as long as the aggregate satisfies the constraint. 
Solving this combinatorial problem exactly is intractable for deep networks, so we adopt a greedy bit-width reduction algorithm that is summarized in \cref{alg:gbmp} and detailed below.

We begin by computing the layer-wise gradients $g_l$ and normalized threshold margin $m_l$ using a calibration set. We define a protection factor
\[
P_l = 1 + \beta(1-m_l),
\]
where $\beta$ controls the influence of the threshold margin. The protection factor increases the cost of reducing the precision of more vulnerable layers, i.e., layers operating closer to the spiking threshold that are more susceptible to quantization-induced perturbations, as explained in \cref{sec:threshold_margin_aware_alloc}. 
We initialize all layers at the maximum candidate precision. At each iteration, we consider reducing the bit-width of every layer to the subsequent lower candidate precision. For each layer, we compute the increase in sensitivity that would result from this reduction and adjust it using the threshold-margin protection factor ($\Delta S_l P_l$). We then divide this value by the number of bits it saves, $n_l \Delta b$; here, \(\Delta b\) is the bit-width step size for the selected transition (e.g., \(\Delta b = 4\) for \(8 \rightarrow 4\) bits) and $n_l$ \(\Delta b\) is the number of bits freed by that reduction. The resulting ratio $\frac{\Delta S_l P_l}{(n_l \Delta b)}$ represents the effective sensitivity increase per bit saved, enabling a fair comparison across layers. The layer with the smallest value is selected, its bit-width is reduced, and the process is repeated until the target average bit-width is reached, as described by
\[
l^* = \arg \min_{l} \frac{\Delta S_l P_l}{n_l \Delta b}, \quad \Delta S_l = S_{l}(b_l - \Delta b) - S_{l}(b_l)
\]
The bit-width of the selected layer is reduced by $\Delta b$, and this process is repeated until the target average bit-width $B$ is reached. Since all sensitivity scores are computed once using a single backward pass (see \cref{alg:gbmp}), the greedy search requires only lightweight score-based comparisons, without any additional gradient evaluations.

\vspace{-0.1em}
\begin{algorithm}[!htt]
\caption{Gradient-Based Mixed-Precision Bit Allocation}
\label{alg:gbmp}
\begin{algorithmic}[1]

\REQUIRE Pretrained network $W$, calibration set $\mathcal D$, candidate bit-widths $\mathcal B$, target bit-width $B_{\text{target}}$
\ENSURE Layer-wise bit allocations $\{b_l\}_{l=1}^{L}$

\STATE Compute layer-wise gradients $\{g_l\}$ using $\mathcal D$
\STATE Compute normalized threshold margins $m_l$ using D
\STATE Compute protection factors $P_l = 1 + \beta(1 - m_l)$

\STATE Initialize all layers with maximum candidate precision

\FOR{each layer $l$}
    \FOR{each candidate bit-width $b \in \mathcal B$}
        \STATE Quantize weights: $W_{q,l}^{(b)} \leftarrow Q(W_l,b)$
        \STATE Compute perturbation: $\Delta W_l^{(b)} = W_{q,l}^{(b)} - W_l$
        \STATE Compute perturbation energy: \\ $\quad \Omega_l^{(b)}=\sum_i g_{l,i}^2(\Delta w_{l,i}^{(b)})^2$
        \STATE Compute normalized sensitivity: \\ $\quad S_l^{(b)}=\Omega_l^{(b)}/(\sum_i g_{l,i}^2+\epsilon)$
    \ENDFOR
\ENDFOR

\WHILE{Average bit-width $> B_{\text{target}}$}
    \FOR{each layer $l$}
        \STATE Compute protected sensitivity increase $\frac{\Delta S_l \, P_l}{n_l \Delta b}$
    \ENDFOR
    \STATE Select the layer with minimum normalized sensitivity score
    \STATE Reduce the selected layer to the next lower candidate bit-width
\ENDWHILE
\RETURN $\{b_l\}_{l=1}^{L}$
\end{algorithmic}
\end{algorithm}

\vspace{-0.3cm}
\section{Experiments and Results}
We evaluate the proposed framework on VGG, ResNet, and ViT-based SNNs using ANN-to-SNN conversion (QCFS~\cite{bu2022optimal}, PASCAL~\cite{ramesh2025pascal}) and direct training (TET~\cite{deng2022tet}) across CIFAR-10/100~\cite{CIFAR}, ImageNet~\cite{imagenet}, and CIFAR10-DVS~\cite{li2017cifar10}. For ViT conversions, we use SpikeZip-TF\cite{pmlr-v235-you24b} with a ViT-Small backbone. All results reported in this section, unless otherwise noted, use post-training quantization directly on SNNs. For all experiments we used a calibration size of $20$, $\beta = 0.7$ and Seed as $42$.
\subsection{Compression Results on SNNs}
\vspace{-0.1em}
\begin{table}[ht]
\centering
\label{tab:quant_results}
\resizebox{\columnwidth}{!}{%
\begin{tabular}{llllcccc}
\toprule
\textbf{Dataset} & \textbf{Architecture} & \textbf{Framework} & \textbf{Method} & \textbf{W-bits} & \textbf{T} & \textbf{W-ratio} & \textbf{Acc. (\%)} \\
\midrule

\multirow{15}{*}{CIFAR-10}
& \multirow{6}{*}{VGG-16}
& \multirow{3}{*}{QCFS-SNN}
& Full Prec. & 32 & 16 & 1.00$\times$ & 95.74 \\
& & & HAWQ-V2 & 4.25 & 16 & 7.54$\times$ & 95.59 \\
& & & \textbf{Ours} & \textbf{4.21} & 16 & \textbf{7.60$\times$} & \textbf{95.64} \\
\cmidrule{3-8}
& & \multirow{3}{*}{PASCAL}
& Full Prec. & 32 & 8 & 1.00$\times$ & 95.58 \\
& & & HAWQ-V2 & 4.01 & 8 & 7.99$\times$ & \textbf{95.19} \\
& & & \textbf{Ours} & \textbf{4.01} & 8 & \textbf{7.99$\times$} & 95.01 \\
\cmidrule{2-8}

& \multirow{6}{*}{ResNet-18}
& \multirow{3}{*}{QCFS-SNN}
& Full Prec. & 32 & 16 & 1.00$\times$ & 94.34 \\
& & & HAWQ-V2 & 4.19 & 16 & 7.63$\times$ & 93.99 \\
& & & \textbf{Ours} & \textbf{4.15} & 16 & \textbf{7.71$\times$} & \textbf{94.01} \\
\cmidrule{3-8}
& & \multirow{3}{*}{PASCAL}
& Full Prec. & 32 & 8 & 1.00$\times$ & 96.50 \\
& & & HAWQ-V2 & 4.00 & 8 & 7.99$\times$ & 96.20 \\
& & & \textbf{Ours} & \textbf{4.01} & 8 & \textbf{7.97$\times$} & \textbf{96.27} \\
\cmidrule{2-8}
& \multirow{3}{*}{SpikeZip-TF}
& \multirow{3}{*}{ANN-SNN}
& Full Prec. & 32 & 64 & 1.00$\times$ & 96.60 \\
& & & HAWQ-V2 & 4.47 & 64 & \textbf{7.16$\times$} & 93.60 \\
& & & \textbf{Ours} & \textbf{4.49} & 64 & 7.12$\times$ & \textbf{94.52} \\
\midrule

\multirow{12}{*}{CIFAR-100}
& \multirow{6}{*}{VGG-16}
& \multirow{3}{*}{QCFS-SNN}
& Full Prec. & 32 & 16 & 1.00$\times$ & 71.67 \\
& & & HAWQ-V2 & 4.46 & 16 & 7.17$\times$ & 71.39 \\
& & & \textbf{Ours} & \textbf{4.39} & 16 & \textbf{7.29$\times$} & \textbf{71.43} \\
\cmidrule{3-8}
& & \multirow{3}{*}{PASCAL}
& Full Prec. & 32 & 8 & 1.00$\times$ & 77.15 \\
& & & HAWQ-V2 & 4.28 & 8 & 7.48$\times$ & \textbf{75.55} \\
& & & \textbf{Ours} & \textbf{4.30} & 8 & \textbf{7.45$\times$} & 73.57 \\
\cmidrule{2-8}

& \multirow{6}{*}{ResNet-18}
& \multirow{3}{*}{QCFS-SNN}
& Full Prec. & 32 & 16 & 1.00$\times$ & 65.74 \\
& & & HAWQ-V2 & 4.20 & 16 & \textbf{7.63$\times$} & 65.51 \\
& & & \textbf{Ours} & \textbf{4.22} & 16 & 7.58$\times$ & \textbf{65.55} \\
\cmidrule{3-8}
& & \multirow{3}{*}{PASCAL}
& Full Prec. & 32 & 8 & 1.00$\times$ & 78.15 \\
& & & HAWQ-V2 & 4.02 & 8 & 7.96$\times$ & 77.01 \\
& & & \textbf{Ours} & \textbf{4.02} & 8 & \textbf{7.96$\times$} & 77.01 \\

\midrule

%\multirow{3}{*}{ImageNet}
%& \multirow{3}{*}{ResNet-34}
%& \multirow{3}{*}{QCFS-SNN}
%& Full Prec. & 32 & 32 & 1.00$\times$ & 74.31 \\
%& & & HAWQ-V2 & 6.64 & 32 & \textbf{4.82$\times$} & 64.96 \\
%& & & \textbf{Ours} & \textbf{6.70} & 32 & 4.78$\times$ & \textbf{67.70} \\
%\cmidrule{3-8}

\multirow{3}{*}{ImageNet}
& \multirow{3}{*}{ResNet-34}
& \multirow{3}{*}{PASCAL}
& Full Prec. & 32 & 8 & 1.00$\times$ & 74.29 \\
& & & HAWQ-V2 & 6.70 & 8 & 4.78$\times$ & 73.14 \\
& & & \textbf{Ours} & \textbf{6.70} & 8 & \textbf{4.78$\times$} & \textbf{73.30} \\

\bottomrule
\end{tabular}
}
\caption{
Comparison of mixed-precision quantization methods on converted SNNs using QCFS~\cite{bu2022optimal} and PASCAL~\cite{ramesh2025pascal}. Results are reported for 4/8-bit layer-wise mixed-precision quantization. W-bits and W-ratio represent the average bit-width and compression ratio, respectively.
}%
% Comparison of quantization methods across conversion-based SNN construction approaches (QCFS-SNN \cite{bu2022qcfs} and PASCAL\cite{ramesh2025pascalpreciseefficientann}).
% The quantization performed is layer wise mixed-precision quantization of 4/8 bits.
% The comparison is done between our own method and HAWQ-V2\cite{dong2020hawqv2}.
% ``W-bits'' denotes average weight bits, ``W-ratio'' denotes the compression ratio, and ``T'' represents the number of simulation timesteps. The quantization algorithm is directly ran on the SNN. 
\label{tab:comparison_conversion}
\end{table}

\vspace{-0.2cm}
\textbf{Conversion-based SNNs.} Table~\ref{tab:comparison_conversion} compares our method with HAWQ-V2 across QCFS, PASCAL, and SpikeZip-TF. Our method achieves comparable or better accuracy while maintaining nearly identical bit-widths and compression ratios. 
On CIFAR-10, it improves SpikeZip-TF by $0.92\%$ (94.52\% vs 93.60\%), while remaining competitive on CIFAR-100. On ImageNet, it slightly outperforms HAWQ-V2 for ResNet34 (73.30\% vs 73.14\%) at same 6.70 bit precision and $4.78\times$ compression.
% On CIFAR-10, the proposed method consistently matches HAWQ-V2 across both QCFS and PASCAL frameworks, and improves the accuracy of SpikeZip-TF by $0.92\%$ (94.52\% vs. 93.60\%). Similar trends are observed on CIFAR-100, where the proposed method maintains competitive performance under the same compression budget, except for PASCAL VGG-16. On ImageNet, our method achieves 73.30\% top-1 accuracy for ResNet-34, slightly outperforming HAWQ-V2 (73.14\%) at the same average weight precision (6.70 bits) and compression ratio (4.78$\times$).

\vspace{0.2cm}
\textbf{Directly trained SNNs.}
Table~\ref{tab:comparison_direct} presents experimental results for SNNs trained using TET. The proposed method consistently produces accuracy comparable to HAWQ-V2, across datasets, while achieving similar average weight precision. In particular, our method improves top-1 accuracy for ResNet-18 and ResNet-20 on CIFAR-100 by 0.22\% and 1.23\%, respectively, while achieving comparable compression ratios. The proposed method also maintains competitive accuracy on the large-scale ImageNet and neuromorphic CIFAR10-DVS datasets using only first-order gradient information for sensitivity estimation.

\begin{table}[!ht]
\centering
\resizebox{\columnwidth}{!}{%
\begin{tabular}{lllccc}
\toprule
\textbf{Dataset} & \textbf{Architecture} & \textbf{Method} & \textbf{W-bits} & \textbf{W-ratio} & \textbf{Acc} (\%)\\
\midrule
\multirow{9}{*}{CIFAR-10}
& \multirow{3}{*}{VGG-16}
& Full Prec. & 32 & 1.00$\times$ & 92.00\\
& & HAWQ-V2 & 4.15 & 7.71$\times$ & 91.92\\
& & \textbf{Ours} & \textbf{4.07} & \textbf{7.87$\times$} & \textbf{91.89}\\
\cmidrule{2-6}
& \multirow{3}{*}{ResNet-18}
& Full Prec. & 32 & 1.00$\times$ & 93.83\\
& & HAWQ-V2 & 4.10 & 7.81$\times$ & 93.74\\
& & \textbf{Ours} & \textbf{4.09} & \textbf{7.82$\times$} & \textbf{93.81}\\
\cmidrule{2-6}
& \multirow{3}{*}{ResNet-20}
& Full Prec. & 32 & 1.00$\times$ & 93.20\\
& & HAWQ-V2 & 4.09 & 7.82$\times$ & 93.02\\
& & \textbf{Ours} & \textbf{4.09} & \textbf{7.82$\times$} & \textbf{93.14}\\
\midrule
\multirow{9}{*}{CIFAR-100}
& \multirow{3}{*}{VGG-16}
& Full Prec. & 32 & 1.00$\times$ & 67.90\\
& & HAWQ-V2 & 4.47 & 7.15$\times$ & 67.83\\
& & \textbf{Ours} & \textbf{4.35} & \textbf{7.36$\times$} & \textbf{67.69}\\
\cmidrule{2-6}
& \multirow{3}{*}{ResNet-18}
& Full Prec. & 32 & 1.00$\times$ & 72.19\\
& & HAWQ-V2 & 4.45 & 7.19$\times$ & 71.56\\
& & \textbf{Ours} & \textbf{4.41} & \textbf{7.26$\times$} & \textbf{71.78}\\
\cmidrule{2-6}
& \multirow{3}{*}{ResNet-20}
& Full Prec. & 32 & 1.00$\times$ & 73.12\\
& & HAWQ-V2 & 4.10 & 7.80$\times$ & 71.59\\
& & \textbf{Ours} & \textbf{4.16} & \textbf{7.68$\times$} & \textbf{72.82}\\
\midrule
\multirow{3}{*}{ImageNet}
& \multirow{3}{*}{ResNet-34}
& Full Prec. & 32 & 1.00$\times$ & 66.70\\
& & HAWQ-V2 & 4.89 & 6.55$\times$ & 65.20\\
& & \textbf{Ours} & \textbf{4.86} & \textbf{6.59$\times$} & \textbf{65.09}\\
\midrule
\multirow{3}{*}{CIFAR10-DVS}
& \multirow{3}{*}{VGGSNN}
& Full Prec. & 32 & 1.00$\times$ & 78.27\\
& & HAWQ-V2 & 4.18 & 7.66$\times$ & 78.26\\
& & \textbf{Ours} & \textbf{4.18} & \textbf{7.66$\times$} & \textbf{78.26}\\
\bottomrule
\end{tabular}%
}
\caption{
Comparison of mixed precision quantization methods on SNNs directly trained using TET~\cite{deng2022tet} for $T{=}6$ timesteps. We use 4/8-bit layer-wise mixed-precision quantization. W-bits and W-ratio indicate the average bit-width and compression ratio, respectively.
}
\vspace{-0.3em}
\label{tab:comparison_direct}
\end{table}

\textbf{Comparison with related works.}
Table~\ref{tab:comparsion_heterogenous} compares our proposed mixed-precision method with the heterogeneous W1/W2 scheme \cite{10.3389/fnins.2025.1665778} and HAWQ \cite{10.1145/3477145.3477158}. The prior W1/W2 method uses 2-bit precision for the first layer and 8-bit for the remaining layers, whereas our method selects 2/4/8-bit widths based on layer-wise quantization sensitivity. We report one configuration on N-MNIST against HAWQ \cite{10.1145/3477145.3477158} and, on CIFAR-10 with VGG-16, improve the compression ratio from $4\times$ to $13.02\times$, while achieving higher accuracy.

\begin{table}[!ht]
\centering
\resizebox{\columnwidth}{!}{%
\begin{tabular}{llcccc}
\toprule
\textbf{Dataset} & \textbf{Architecture} & \textbf{Method} & \textbf{W-bits} & \textbf{W-ratio} & \textbf{Acc} (\%)\\
\midrule
\multirow{2}{*}{CIFAR-10} & \multirow{2}{*}{VGG-16} & Two-level & 2/8 & 4$\times$ & 91.55 \\
& & \textbf{Ours} & \textbf{2.46} & \textbf{13.02$\times$} & \textbf{91.88} \\
\midrule
\multirow{2}{*}{CIFAR-100} & \multirow{2}{*}{ResNet-19} & Two-level & 2/8 & 4$\times$ & 72.23 \\
& & \textbf{Ours} & \textbf{2.89} & \textbf{11.07$\times$} & \textbf{72.00} \\
\midrule
\multirow{2}{*}{N-MNIST} & \multirow{2}{*}{3-layer CNN} & Hessian & 5.33 & 5.97$\times$ & 93.6 \\
& & \textbf{Ours} & \textbf{4.02} & \textbf{7.96$\times$} & \textbf{93.5} \\
\bottomrule
\end{tabular}%
}
\caption{
Comparison of our quantization method extended to perform 2/4/8-bit mixed precision with the heterogeneous two-level precision scheme $(W1/W2)$ proposed in \cite{10.3389/fnins.2025.1665778} and HAWQ \cite{10.1145/3477145.3477158}.
}
%\vspace{-1em}
\label{tab:comparsion_heterogenous}
\end{table}

\begin{figure}[!ht]
    \centering
    \includegraphics[width=0.44\textwidth]{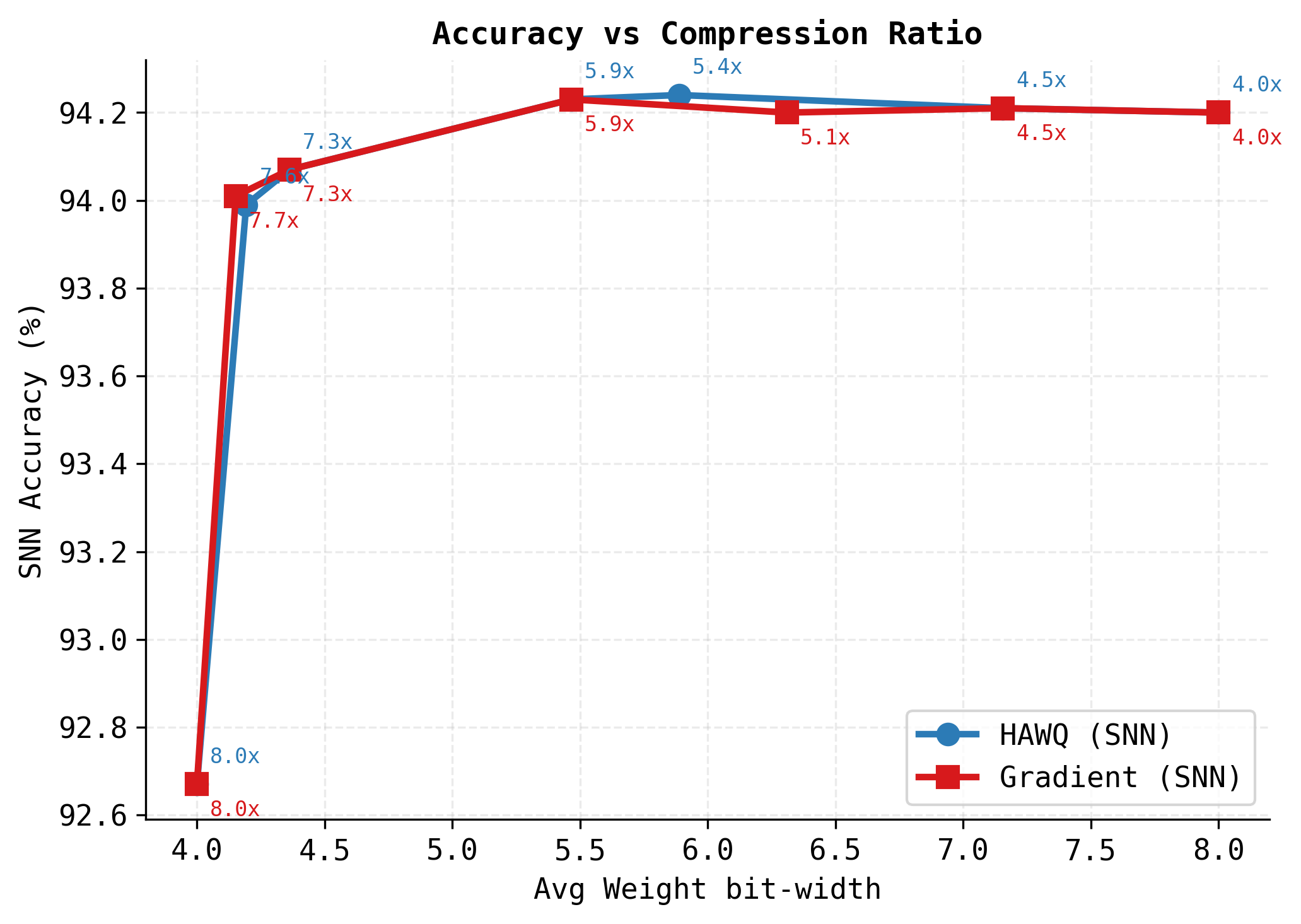}
    \caption{Post-training quantization compression ratio achieved by HAWQ-V2 \cite{NEURIPS2020_d77c7035} and the proposed method on ResNet-18 trained on CIFAR-10 using a 4/8-bit mixed-precision configuration.}
    \vspace{-0.1em}
    \label{fig:compression}
\end{figure}

\vspace{0.2cm}
\textbf{Accuracy vs.\ Compression Trade-off.}
As illustrated in Figure~\ref{fig:compression}, our method achieves accuracy-compression trade-off comparable to that provided by HAWQ-V2~\cite{NEURIPS2020_d77c7035} across all average bit-width configurations on ResNet-18 trained on CIFAR-10. At aggressive compression ratios (7.6--7.7$\times$, corresponding to $\sim$4.1-bit average precision), both methods incur a modest accuracy drop to approximately 94.0\%, recovering to 94.2\% as the average bit-width increases beyond 5.5 bits. Overall, the two curves remain within 0.05\% of each other for the entire bit-width range, confirming that first-order gradient sensitivity serves as effective and computationally cheaper alternative for Hessian-based quantization sensitivity in SNNs.

\vspace{-0.3cm}
\subsection{Computation and Efficiency}
\begin{figure}[!ht]
    \centering
    \includegraphics[width=\columnwidth]{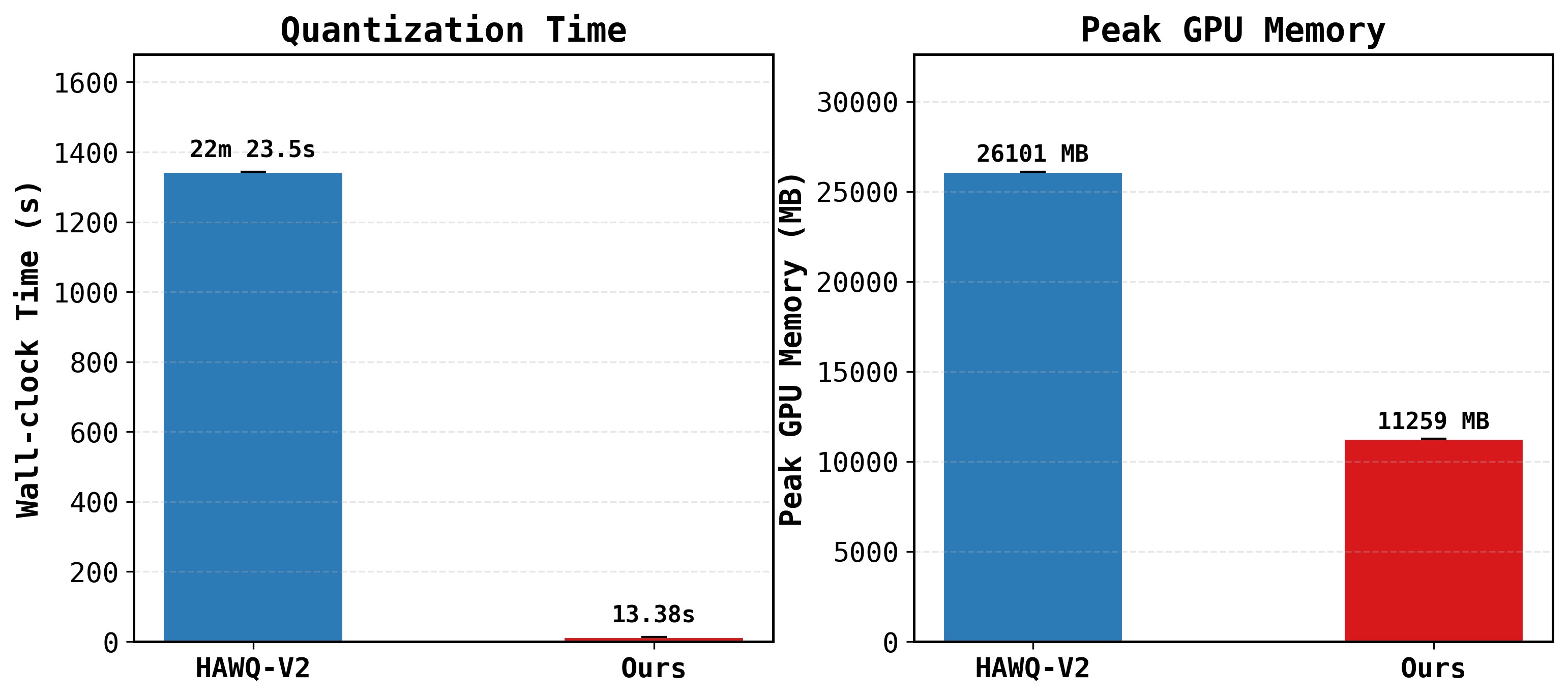}
    \caption{GPU memory needed (MB) and total quantization time taken by HAWQ-V2 vs. our method to quantize ResNet-34 SNN on the ImageNet dataset, measured on ADA6000 GPU.}
    \vspace{-0.3em}
    \label{fig:gpu_mem_and_time}
\end{figure}

Figure~\ref{fig:gpu_mem_and_time} illustrates that the proposed method is significantly more memory- and time-efficient than HAWQ-V2 during bit allocation. The higher cost of HAWQ-V2 arises from repeated Hutchinson trace estimation for Hessian approximation, which is further amplified by temporal unrolling in SNNs. By relying on a single first-order backward pass to estimate layer sensitivities, the proposed method eliminates this overhead, resulting in substantially lower memory consumption and runtime. Specifically, on ResNet-34 trained on the ImageNet dataset, the proposed method reduces peak GPU memory from 26.1 GB to 11.3 GB (i.e. \textbf{56.9\% reduction}) and decreases bit-allocation time from 22 min 23.5 s to 13.38 s, achieving an approximately \textbf{100× speedup} over HAWQ-V2.

\begin{figure*}[t]
    \centering
    \includegraphics[width=\textwidth]{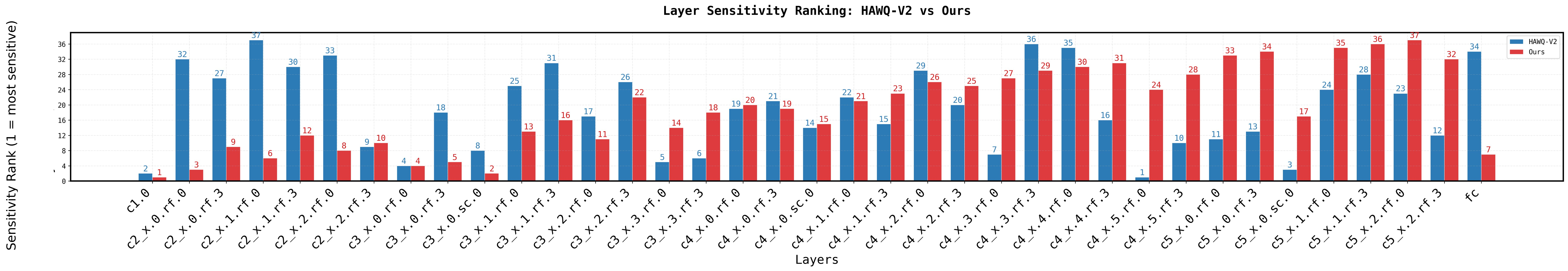}
    \caption{Sensitivity-wise layer ranking of ResNet34 SNN trained on ImageNet and quantized using HAWQ-V2~\cite{NEURIPS2020_d77c7035} vs. our method. Note that c1 denotes the initial conv layer, while labels of the form cX.x.B.rf/sc.K denote the K-th convolution in the residual (rf) or shortcut (sc) branch of the B-th residual block in the convX-x stage.}
    \vspace{-0.3em}
    \label{fig:sensitivity_each_layer}
\end{figure*}

\begin{figure*}[t]
    \centering
    \includegraphics[width=\textwidth]{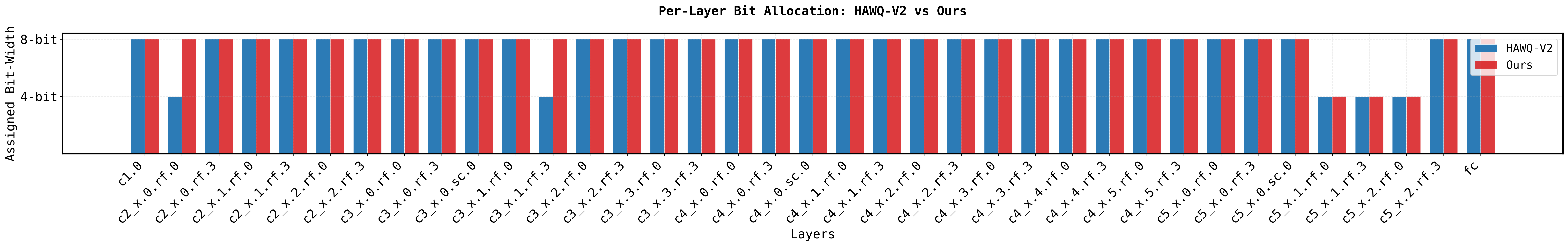}
    \caption{Layer-wise mixed precision bit allocation of HAWQ-V2~\cite{NEURIPS2020_d77c7035} versus our method for ResNet-34 SNN on ImageNet. Note that c1 denotes the initial conv layer, while labels of the form cX.x.B.rf/sc.K denote the K-th convolution in the residual (rf) or shortcut (sc) branch of the B-th residual block in the convX-x stage.}
    \vspace{-0.3em}
    \label{fig:bit_allocation}
\end{figure*}
% \subsection{Sensitivity and Bit Allocation Analysis}
% In the Figure~\ref{fig:sensitivity_each_layer}, we can see how individual layers are ranked by HAWQ-V2 and our method for the ResNet-18 SNN model. while \cref{fig:bit_allocation}
%  shows the corresponding 4/8-bit mixed-precision assignment. The sensitivity ranking differe across several layers because HAWQ-V2 estimates layer importance using second-order Hessian information, whereas our method relies on first-order loss gradients. Consequently, the two approaches assign different relative importance to many intermediate layers. However, these differences have little impact on the final mixed-precision configuration. As shown in \cref{fig:bit_allocation}, both methods produce nearly identical bit allocations, with only three layers receiving different bit-width assignments. This close agreement of bit allocations demonstrates that the proposed first-order sensitivity metric effectively identifies the layers requiring higher precision, producing a mixed-precision configuration comparable to HAWQ-V2 while avoiding the computational overhead associated with Hessian-based sensitivity estimation.

\subsection{Sensitivity and Bit Allocation Analysis}
Figures~\ref{fig:sensitivity_each_layer} and~\ref{fig:bit_allocation} compare the layer ranking and the resulting 4/8-bit assignment of HAWQ-V2 and the our method for ResNet-18. Although the rankings differ for some intermediate layers due to the use of Hessian versus gradient-based sensitivity, the resulting bit allocations are highly consistent, with only two layers receiving different precisions. This demonstrates that our first-order sensitivity can achieve a comparable mixed-precision configuration without the computational overhead of Hessian estimation. Additional analysis is provided in the supplementary material.

\subsection{Post-Quantization Fine-Tuning Analysis}
Although the proposed framework is primarily designed for post-training quantization (PTQ), it can be extended to perform post-quantization fine-tuning similar to HAWQ-V2. As shown in Table~\ref{tab:comparison_direct_extreme}, the proposed bit allocation remains effective under aggressive 2/4/8-bit mixed-precision quantization, achieving over $12\times$ compression while maintaining accuracy close to full-precision models.

\begin{table}[!ht]
\centering
\resizebox{\columnwidth}{!}{%
\begin{tabular}{llcccc}
\toprule
\textbf{Dataset} & \textbf{Architecture} & \textbf{Method} & \textbf{W-bits} & \textbf{W-ratio} & \textbf{Acc} (\%)\\
\midrule
\multirow{4}{*}{CIFAR-10}
& \multirow{2}{*}{VGG-16}
& Full Prec. & 32 & 1.00$\times$ & 92.00\\
& & \textbf{Ours} & \textbf{2.46} & \textbf{13.02$\times$} & \textbf{91.88}\\
\cmidrule{2-6}
& \multirow{2}{*}{ResNet-18}
& Full Prec. & 32 & 1.00$\times$ & 93.83\\
& & \textbf{Ours} & \textbf{2.52} & \textbf{12.68$\times$} & \textbf{93.39}\\
\midrule
\multirow{4}{*}{CIFAR-100}
& \multirow{2}{*}{VGG-16}
& Full Prec. & 32 & 1.00$\times$ & 67.90\\
& & \textbf{Ours} & \textbf{2.47} & \textbf{12.98$\times$} & \textbf{66.41}\\
\cmidrule{2-6}
& \multirow{2}{*}{ResNet-18}
& Full Prec. & 32 & 1.00$\times$ & 72.19\\
& & \textbf{Ours} & \textbf{2.43} & \textbf{13.19$\times$} & \textbf{71.03}\\

\midrule

\multirow{2}{*}{ImageNet}
& \multirow{2}{*}{ResNet-34}
& Full Prec. & 32 & 1.00$\times$ & 66.70\\
& & \textbf{Ours} & \textbf{2.58} & \textbf{12.37$\times$} & \textbf{64.12}\\

\bottomrule
\end{tabular}%
}
\caption{
Comparison of our method extended to perform 2/4/8-bit mixed-precision quantization, on SNNs trained using TET~\cite{deng2022tet} for $T{=}6$ timesteps, versus full-precision baselines. W-bits and W-ratio indicate the average weight precision and compression ratio, respectively.
}
\label{tab:comparison_direct_extreme}
\end{table}

\subsection{Ablation Study}
We conduct ablation studies to evaluate three aspects of the proposed framework: (i) its applicability to conventional ANN models, (ii) the contribution of threshold-margin information to mixed-precision bit allocation, and (iii) its robustness across different surrogate gradient functions.

\subsubsection{Generalization to ANN Models} 
Although the proposed framework is motivated by efficient SNN quantization, its gradient-based sensitivity metric is applicable to standard ANNs. Table~\ref{tab:ann_comparison} compares our method with HAWQ-V2 on CIFAR-10 and ImageNet. Since ANNs do not exhibit spike dynamics, only the first-order gradient-based sensitivity metric (without threshold-margin information) is used. Using candidate bit-widths of \{1,2,4,8\}, the proposed method achieves higher compression ratios and better accuracy than HAWQ-V2. On CIFAR-10 and ImageNet, the proposed method improves the compression ratio by 7.6\% and 13.0\%, respectively, while increasing top-1 accuracy by 0.39\% and 0.43\%, respectively, over HAWQ-V2. These results indicate that the proposed first-order sensitivity metric remains effective beyond SNNs.%, providing an efficient alternative to Hessian-based sensitivity estimation.

\begin{table}[ht]
\centering
\resizebox{\columnwidth}{!}{%
\begin{tabular}{lllllccccc}
\toprule
\textbf{Dataset} & \textbf{Arch} & \textbf{Type} & \textbf{Method} & \textbf{W-bits} & \textbf{W-ratio} & \textbf{Acc} (\%)\\
\midrule
\multirow{3}{*}{CIFAR-10}
  & \multirow{3}{*}{ResNet20} 
  & \multirow{3}{*}{ANN}
  & Full Prec. & 32 & 1.00$\times$ & 92.90\\
  & & & HAWQ-V2 & $2_{MP}$ & 13.11$\times$ & 92.22\\
  & & & \textbf{Ours} & \textbf{2.27} & \textbf{14.11$\times$} & \textbf{92.61}\\
\midrule
\multirow{3}{*}{ImageNet}
  & \multirow{3}{*}{Inception-V3} 
  & \multirow{3}{*}{ANN}
  & Full Prec. & 32 & $1.00\times$ & 77.45\\
  & & & HAWQ-V2 & $2_{MP}$ & $12.04\times$ & 75.68\\
  & & & \textbf{Ours} & \textbf{2.35} & \textbf{13.61$\times$} & \textbf{76.11}\\
\bottomrule
\end{tabular}%
}
\caption{
Comparison of our method on ANNs with HAWQ-V2 whose results are taken from~\cite{NEURIPS2020_d77c7035}. Our method uses candidate bit-widths of {1, 2, 4, 8}. For HAWQ-V2, the reported ``W-bits'' correspond to the minimum candidate bit-width, whereas for our method, ``W-bits'' indicate the average allocated weight precision. 
}
\label{tab:ann_comparison}
\end{table}
\subsubsection{Effect of Threshold Margin}
Table~\ref{tab:ablation_spike} evaluates the contribution of threshold-margin information to the proposed sensitivity metric. Compared with gradients alone, integrating threshold-margin information consistently improves quantized accuracy, with a particularly large gain for CIFAR-100 (VGG-16) from 57.28\% to 69.94\% at  approximately 2.5 bits. These experiments use $\beta=0.7$ and candidate bit-widths of $\{1,2,\ldots,8\}$ , allowing the bit-allocation algorithm to exploit a wider range of precision levels. Further analysis of the effect of $\beta$ is provided in the supplementary material.
% The results indicate that threshold-margin information complements gradient-based sensitivity estimation by better identifying layers that require higher precision.

\begin{table}[ht]
\centering
\resizebox{\columnwidth}{!}{%
\begin{tabular}{ll ccc ccc}
\toprule
\multirow{2}{*}{\textbf{Dataset}} & \multirow{2}{*}{\textbf{Architecture}}
& \multicolumn{3}{c}{\textbf{Grad}}
& \multicolumn{3}{c}{\textbf{Grad + Threshold Margin}} \\
\cmidrule(lr){3-5} \cmidrule(lr){6-8}
& & \textbf{W-bits} & \textbf{W-ratio} & \textbf{Acc (\%)}
  & \textbf{W-bits} & \textbf{W-ratio} & \textbf{Acc (\%)} \\
\midrule
\multirow{3}{*}{CIFAR-10}
& ResNet-18 & 2.72 & 11.73$\times$ & 93.72 & 2.76 & 11.56$\times$ & \textbf{94.24} \\
\cmidrule{2-8}
& \multirow{2}{*}{VGG-16}
  & 2.74 & 11.64$\times$ & 95.55 & 2.74 & 11.64$\times$ & \textbf{95.61} \\
& & 2.49 & 12.83$\times$ & 95.46 & 2.46 & 13.01$\times$ & \textbf{95.49} \\
\midrule
\multirow{5}{*}{CIFAR-100}
& \multirow{2}{*}{ResNet-18}
  & 3.63 & 8.81$\times$ & 64.24 & 3.72 & 8.60$\times$ & \textbf{64.71} \\
& & 3.56 & 8.98$\times$ & 64.24 & 3.59 & 8.89$\times$ & \textbf{64.6} \\
\cmidrule{2-8}
& \multirow{3}{*}{VGG-16}
  & 2.99 & 10.72$\times$ & 70.76 & 2.96 & 10.78$\times$ & \textbf{70.85} \\
& & 2.64 & 12.11$\times$ & 70.22 & 2.65 & 12.09$\times$ & \textbf{70.54} \\
& & 2.44 & 13.07$\times$ & 57.28 & 2.48 & 12.89$\times$ & \textbf{69.94}    \\
\bottomrule
\end{tabular}%
}
\caption{Ablation to study the impact of threshold-margin information in the proposed sensitivity metric. Results employ candidate bit-widths of $\{1,2,\ldots,8\}$ in the QCFS framework ($T=16$).
}
\vspace{-0.3em}
\label{tab:ablation_spike}
\end{table}

\subsubsection{Robustness across Surrogate Gradients}
\begin{figure}[!ht]    
    \includegraphics[width=\columnwidth]{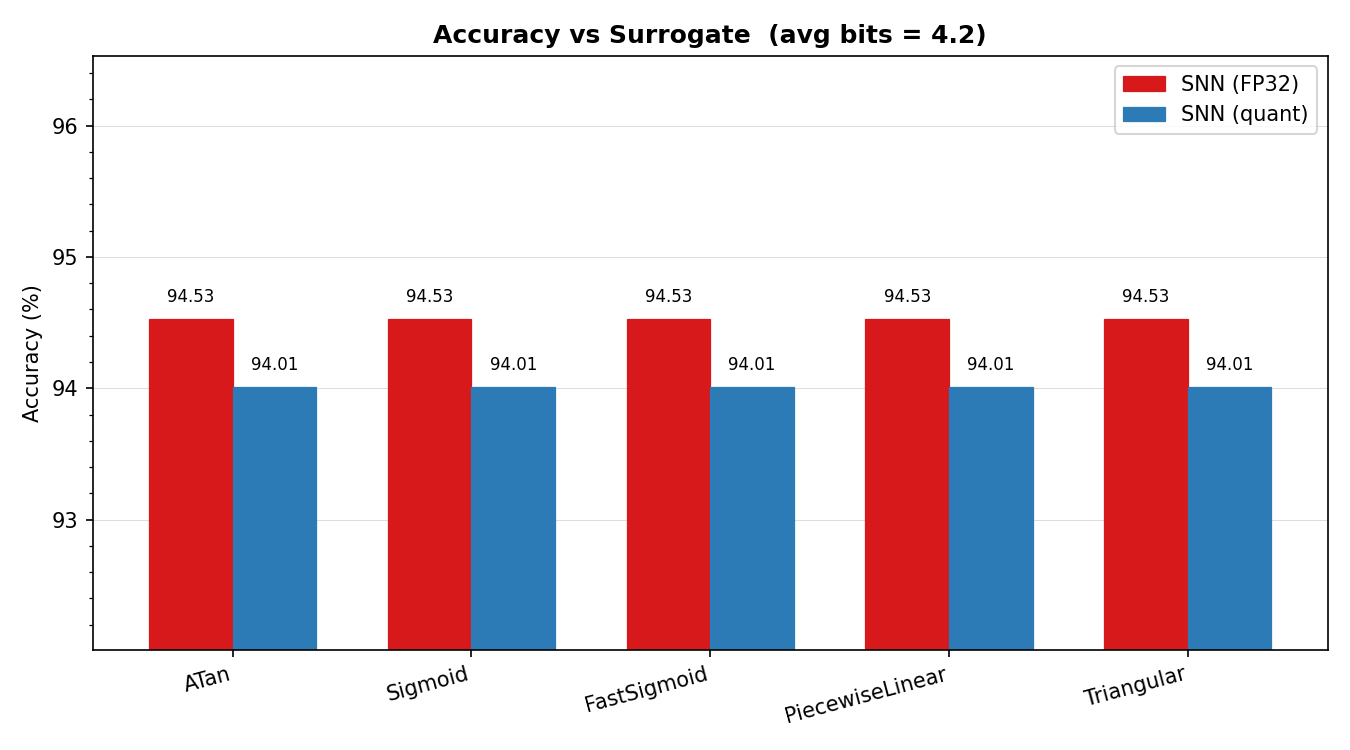}
    \caption{Accuracy of ResNet-18 on CIFAR-10 dataset using different surrogate gradients}
    \label{fig:surrogate_accuracy}
\end{figure}

\begin{figure}[!ht]    
    \includegraphics[width=\columnwidth]{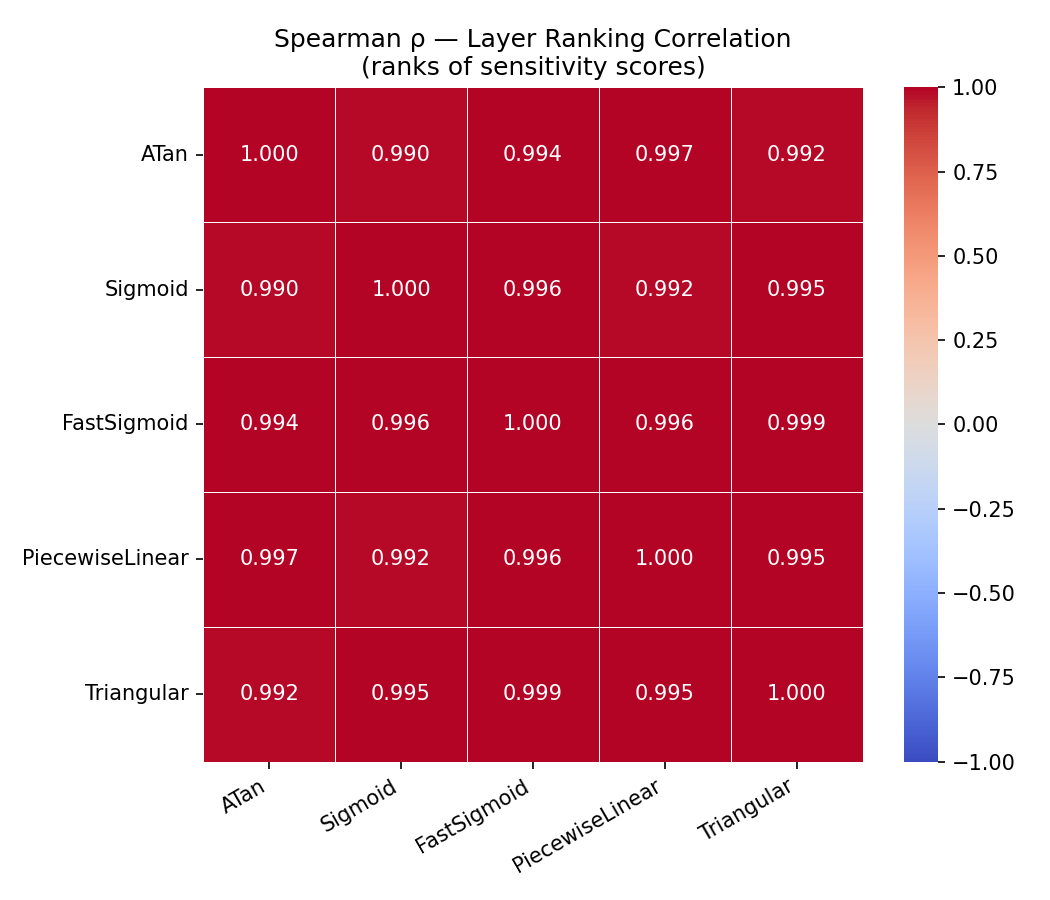}
    \caption{Pairwise Spearman rank correlation~\cite{spearman1904} between layer sensitivity rankings obtained using different surrogate gradient functions. High correlation indicates that the proposed sensitivity metric produces nearly identical layer orderings across surrogate gradient functions.}
    \label{fig:surrogate_correlation}
\end{figure}

Since gradients in SNNs depend on surrogate function, we evaluate the proposed framework using five commonly used surrogate gradients while keeping the trained model and quantization settings fixed. As shown in Figure~\ref{fig:surrogate_accuracy}, the proposed method achieves nearly identical quantized accuracy across all surrogate functions. Moreover, Figure~\ref{fig:surrogate_correlation} shows that the resulting layer sensitivity rankings are highly consistent, with pairwise Spearman coefficients~\cite{spearman1904} exceeding 0.99. Thus, although surrogate functions may change the absolute sensitivity values, they preserve the relative ordering of layers, leading to consistent bit-allocation decisions. Additional analysis are provided in supplementary materials.

\vspace{-0.2cm}
\subsection{Hardware Efficiency: ANN vs.\ SNN}
\label{sec:hw_ann_snn}
\begin{figure}[!ht]
    \centering
    \includegraphics[width=0.5\textwidth]{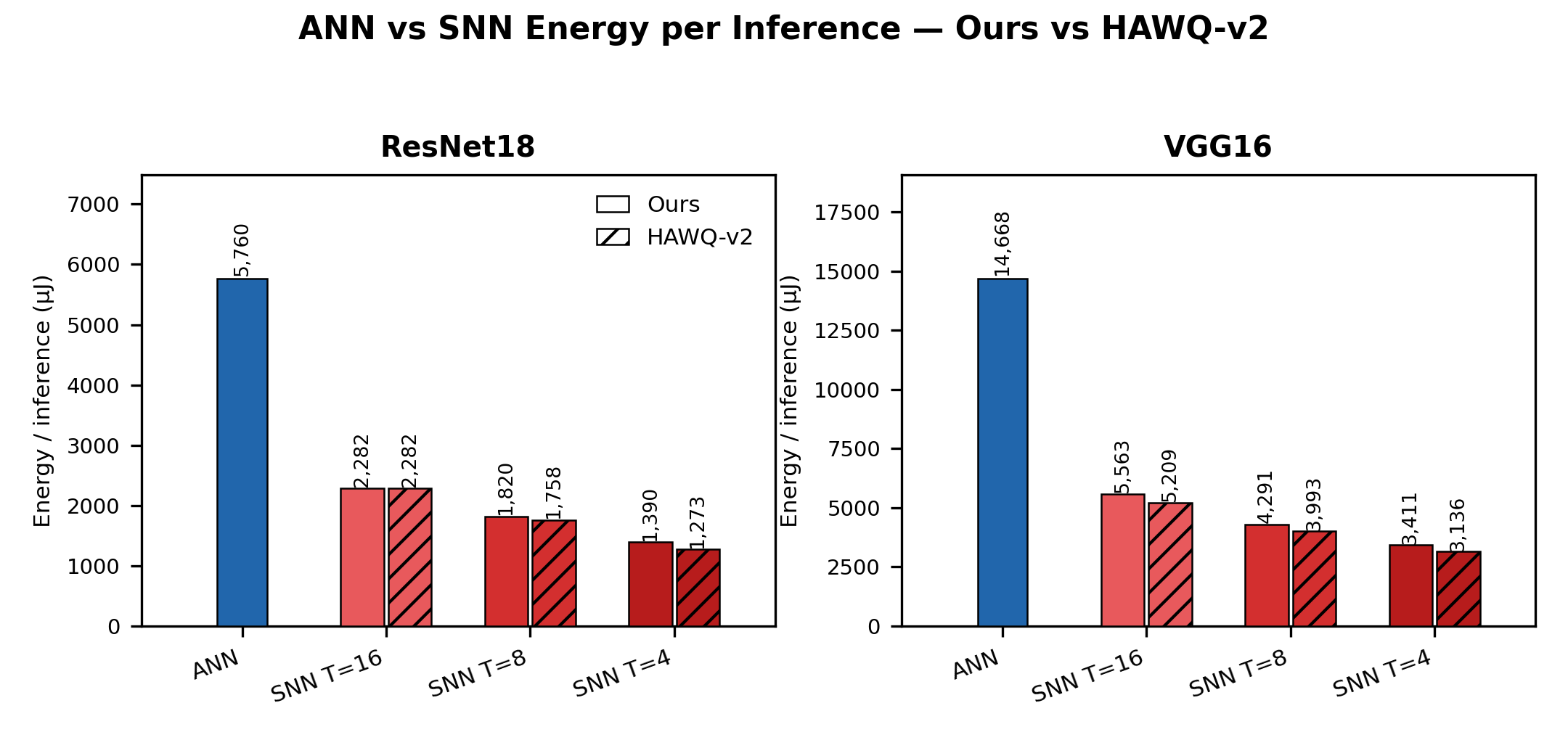}
    \caption{Per-inference energy: SparTen-style ANN baseline vs. LoAS-style SNN baseline for ResNet-18 and VGG-16.}
    \label{fig:hw_ann_snn}
    \vspace{-0.3em}
\end{figure}

\begin{figure}[!ht]
    \centering
    \includegraphics[width=0.5\textwidth]{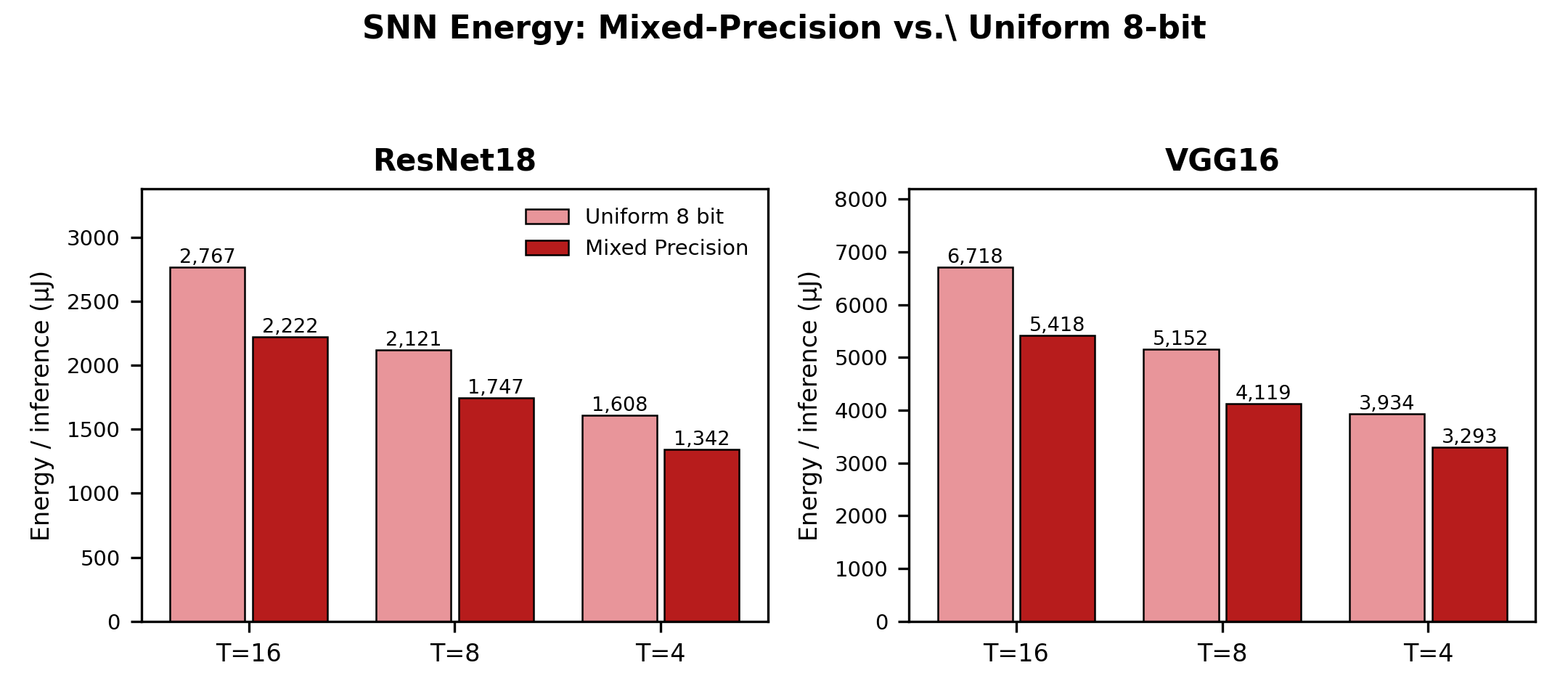}
    \caption{Energy incurred by ResNet-18 SNN and VGG-16 SNN per-inference for mixed-precision versus uniform 8-bit weights.}
    \label{fig:snn_mixed_vs_uniform}
    \vspace{-0.3em}
\end{figure}
We simulate VGG-16 and ResNet-18 on SparTen~\cite{sparten}, an ANN hardware accelerator, against LoAS~\cite{loas}, an SNN hardware accelerator. Both accelerators use the same dual-sparse processing element (PE) pipeline and operate at 560\, MHz. Figure~\ref{fig:hw_ann_snn} illustrates per-inference energy at $T \in \{4,8,16\}$. 
The SNN baseline consumes $2.6$--$2.7\times$ less energy than the ANN baseline at $T{=}16$, rising to $4.3$--$4.5\times$ at $T{=}4$, driven by both lower per-PE power and higher combined sparsity at smaller timesteps. %This gain comes with an accuracy cost at low $T$, motivating accuracy-aware quantization at moderate timestep budgets where SNNs already hold a clear energy edge over ANNs.
Both HAWQ-V2 and our method yield mixed-precision SNNs that provide comparable energy efficiency. Within the SNN baseline itself, we further compare mixed-precision (4/8-bit) weights against uniform 8-bit weights at $T \in \{4,8,16\}$, as shown in Figure \ref{fig:snn_mixed_vs_uniform}. 
Mixed-precision weights consistently reduce SNN energy by $19$--$25\%$ over uniform 8-bit execution across all $T$, for both architectures, since the lower average bit-width directly reduces per-PE dynamic power.

\section{Conclusion}
In this work, we proposed a gradient-based mixed-precision quantization framework for SNNs that avoids costly Hessian-based sensitivity estimation through an efficient first-order, spike-aware formulation. By incorporating membrane threshold information into bit allocation, our method captures SNN-specific temporal dynamics while requiring only a single backward pass, achieving competitive accuracy with reduced computational and memory costs for resource-constrained neuromorphic deployment.

%%%%%%%%% REFERENCES
{\small
\bibliographystyle{ieee_fullname}
\bibliography{references}
}

\clearpage
\setcounter{page}{1}          % optional: restart page numbers for the supplementary
\maketitlesupplementary       % provided by the ICCV/CVPR-style kits; see note below

% Restart numbering so "Sec. 1 of the supplementary" in the main text is correct
\setcounter{section}{0}
\setcounter{figure}{0}
\setcounter{table}{0}
\setcounter{equation}{0}
\renewcommand{\thesection}{S\arabic{section}}
\renewcommand{\thefigure}{S\arabic{figure}}
\renewcommand{\thetable}{S\arabic{table}}
\section{SNN Training Paradigms}

SNNs process information through discrete spiking events over multiple timsteps. At each timestep $t = 1\dots, T$, a spiking neuron integrates incoming weighted spikes into its membrane potential. When the membrane potential exceeds a predefined firing threshold $V_{th}$, the neuron emits a binary spike and updates its internal state. The event-driven computation leads to sparse neural activity and enables energy-efficient inference on neuromorphic hardware.

Since SNNs produce outputs over multiple timesteps, the final prediction is typically obtained by aggregating the output spike across the entire simulation window. A commonly used decoding strategy is rate encoding, where the predicted class is computed as 
\[
\hat{y} = \arg\max_{c} \left(\frac{1}{T} \sum_{t=1}^{T} O_c(t) \right),
\]
where $O_c(t)$ denotes the output spike activity corresponding to class $c$ at timestep $t$. Broadly, SNNs can be trained using two different paradigms: direct training, which optimizes the network using surrogate gradients to overcome the non-differentiability of spike functions, and ANN-to-SNN conversion, which converts a pre-trained ANN into an equivalent SNN while preserving its predictive performance.

\paragraph{\textbf{Direct Training of SNN using TET ~\cite{deng2022tet}.}} 
Temporal Efficient Training improves the optimization of directly trained SNNs by applying supervision across all timesteps rather than only at the final timestep. Instead of computing the loss solely from the final output, TET computes the average temporal loss over the entire spike sequence:
\[ 
\mathcal{L}_{TET} = \frac{1}{T} \sum_{t=1}^{T} \mathcal{L}_{CE}(O(t), y),
\]
which prevents gradient collapse in early timestep and significantly improves accuracy over standard rate-coded training. However, the aggregation of surrogate gradients across time introduces new sensitivity patterns that affect adversarial bit-level saliency, as shown in our experiments.
\paragraph{\textbf{ANN-to-SNN Conversion using QCFS~\cite{bu2022optimal}.}} QCFS is an ANN-to-SNN conversion framework designed to achieve accurate and low-latency spiking inference. Instead of using the ReLU activation during ANN training, it uses the \textbf{Quantization Clip Floor-Shift\;(QCFS)} activation function. This activation constrains the network output to discrete levels that closely resemble the spike rates produced by SNNs, thereby reducing the mismatch between ANN activations and SNN firing behavior. 
During conversion, the trained ANN is transformed into an equivalent SNN by replacing the quantized activations with spiking neurons while preserving the learned thresholds and scaling factors. As a result, the converted SNN achieves high accuracy using fewer simulation timesteps, making QCFS an efficient ANN-to-SNN conversion framework.
\paragraph{\textbf{Transformer-based SNN.}} To demonstrate the generality of the proposed mixed-precision quantization framework, we evaluate our method on spike-based vision Transformer (ViT) models using the SpikeZIP-TF \cite{pmlr-v235-you24b} ANN-to-SNN conversion framework. Unlike convolutional networks, Vision Transformers rely on self-attention, softmax, and layer normalization operations, which are not directly compatible with spike-based computation.

SpikeZIP-TF addresses this issue by first quantizing a pretrained Vision Transformer through quantization-aware training. During conversion, ANN activation is replaced by a spiking neuron called ST-BIF$^+$. SpikeZIP-TF proposes \textbf{Spike-Equivalent Self-Attention (SESA)}, \textbf{Spike-Softmax} and \textbf{Spike-LayerNorm}, which preserve the functionality of the corresponding Transformer modules in the spiking domain. These operators enable the converted SNN to closely mimic the behavior of the quantized Vision Transformer models, while maintaining high classification accuracy with relatively fewer simulation timesteps. 

\section{Theoretical Justification of the First-Order Sensitivity Metric}

Hessian-based methods estimate parameter sensitivity using the curvature of the loss landscape. For a model with parameters $w$, the Hessian matrix is defined as 
\[
H = \nabla^2_w L(w),
\]
where $L(w)$ denotes the training loss. The Hessian captures the local curvature. Computing Hessian is expensive, particularly for temporally unrolled spiking neural networks.
We instead utilize first-order gradient statistics derived from the Fisher information matrix~\cite{3019383,JMLR:v21:17-678}, which measures the sensitivity of the model output distribution with respect to parameter perturbations. 
% Parameters associated with larger Fisher values produce greater changes in the network predictions under small perturbations, indicating higher importance and quantization sensitivity.

Consider a probabilistic model that predicts a conditional distribution:
\[
p(y|x;w),
\]
where $w$ are the parameters of this model, $x$ are the input data, and $y$ are the model predictions. Under the negative log-likelihood training, the loss is defined as:
\[
L(w) = -\log p(y|x;w).
\]
To measure how sensitive the output distribution is to parameter changes, we compute the gradient of the log-likelihood with respect to the parameters:
\[
s(w) = \nabla_w \log(y|x;w)
\]
Here, $s(w)$ is known as the score function. While the score function measures the sensitivity of the model distribution for an individual sample, quantization sensitivity should capture the expected parameter sensitivity over the entire data distribution. Therefore, we consider the expected covariance of the score function, leading to the Fisher information matrix~\cite{JMLR:v21:17-678, 3019383}:
\[
F = \mathbb{E}[s(w)s(w)^T].
\]
The Fisher information matrix~\cite{JMLR:v21:17-678,3019383} measures the expected sensitivity of the model distribution over the data distribution. Parameters associated with larger Fisher values produce greater changes in the network predictions under small perturbations, indicating higher importance and quantization sensitivity, thus Fisher captures parameter importance. 
Since the training objective is defined using negative log-likelihood,
the gradient of the loss with respect to the parameters can be written as 
\[
g = \nabla_wL(w) = -\nabla_w\log p(y|x;w).
\]
Using the definition of the score function,
\[
s(w) = \nabla_w\log p(y|x;w),
\]
we obtain
\[
g = -s(w).
\]
Now, substituting this into the Fisher information matrix~\cite{JMLR:v21:17-678} gives
\[
F = \mathbb{E}{[gg^T]}.
\]
The diagonal entries of the Fisher matrix satisfy
\[
F_{ii} = \mathbb{E}{[g_i^2]},
\]
where $g_i$ denotes the gradient corresponding to parameter $w_i$, formulated as
\[
    g_i = \frac{\partial L}{\partial w_i}.
\]
Parameters with larger squared gradients produce larger changes in the loss under perturbations and are therefore more sensitive to quantization.
Since
\[
L(w) = -\log p(y|x;w),
\]
the Hessian of the loss is
\[
H = \nabla_w^2L(w) = -\nabla_w^2\log p(y|x;w).
\]
Under negative log-likelihood objectives, the curvatures of the loss can be related to the statistical properties of the underlying probability distribution. The relationship can be derived from the normalized property of probability distributions, described by
\[
\int p(y|x;w)dy = 1.
\]
Differentiating the normalization condition with respect to the model parameters gives
\[
\nabla_w \int p(y|x;w)dy = 0,
\]
\[
\int \nabla_w p(y|x;w) = 0.
\]
Using the identity~\cite{bishop:2006:PRML}:
\[
\nabla_w p(y|x;w) = p(y|x;w)\nabla_w\log p(y|x;w),
\]
the above expression becomes
\[
\int p(y|x;w)\nabla_w\log p(y|x;w)dy = 0.
\]
Therefore,
\[
\mathbb{E}[s(w)] = 0.
\]
The above expression is known as the score-function identity, which states that the expected score over the data distribution is zero.
Now, differentiating once again with respect to the model parameters and rearranging yields
\[
-\mathbb{E}[\nabla_ws(w)] = \mathbb{E}[s(w)s(w)^T].
\]
The right-hand side corresponds to the Fisher Information matrix, while the left-hand side represents the expected Hessian curvature of the negative log-likelihood objective~\cite{JMLR:v21:17-678,3019383}. 
Therefore,
\[
F = \mathbb{E}[ht].
\]
For models trained near local optima, the empirical Hessian closely approximates its expectation over the data distribution, yielding
\[
F \approx H;
\]
\[
g_i^2 = H_{ii},
\]
allowing the second-order loss perturbation to be efficiently approximated using squared gradients, formulated as
\[
\Delta L \approx \frac{1}{2}\sum_i g_i^2 (\Delta w_i)^2.
\]

\section{Additional Experiments and Comparison with QP-SNN \texorpdfstring{\cite{wei2025qpsnn}}{[Wei et al., 2025]}}

Table~\ref{tab:comparison_qpsnn} compares our mixed-precision framework with the uniform quantization and pruning strategy adopted by QP-SNN \cite{wei2025qpsnn}. At comparable compression levels, our proposed method consistently achieves higher accuracy across all evaluated datasets. Notably on ImageNet, the proposed method achieves 65.09\% compared to 61.36\% and 58.05\% for the 8-bit and 4-bit QP-SNN models, respectively. These gains highlight the effectiveness of mixed-precision over uniform-bit allocation.

\begin{table}[ht]
\centering
\resizebox{\columnwidth}{!}{%
\begin{tabular}{llcccc}
\toprule
\textbf{Dataset} & \textbf{Architecture} & \textbf{Method} & \textbf{W-bits} & \textbf{W-ratio} & \textbf{Acc} (\%)\\
\midrule
\multirow{3}{*}{CIFAR-10} & \multirow{3}{*}{VGG-16} & QP-SNN & 8 & 4$\times$ & 91.98 \\
& & QP-SNN & 4 & 8$\times$ & 91.90 \\
& & \textbf{Ours} & \textbf{4.07} & \textbf{7.87$\times$} & \textbf{91.89} \\
\midrule
\multirow{3}{*}{CIFAR-100} & \multirow{3}{*}{VGG-16} & QP-SNN & 8 & 4$\times$ & 66.69 \\
& & QP-SNN & 4 & 8$\times$ & 66.21 \\
& & \textbf{Ours} & \textbf{4.35} & \textbf{7.36$\times$} & \textbf{67.69} \\
\midrule
\multirow{3}{*}{ImageNet} & \multirow{3}{*}{ResNet-34} & QP-SNN & 8 & 4$\times$ & 61.36 \\
& & QP-SNN & 4 & 8$\times$ & 58.06 \\
& & \textbf{Ours} & \textbf{4.86} & \textbf{6.59$\times$} & \textbf{65.09} \\
\bottomrule
\end{tabular}%
}
\caption{
Comparison of our mixed-precision quantization method with QP-SNN that uses uniform 8-bit and 4-bit quantization.
}
\label{tab:comparison_qpsnn}
\end{table}

\section{Computational Overhead of Sensitivity Estimation}

Let \(L\) denote the number of quantizable layers, \(N=\sum_l n_l\) the
total number of parameters, \(K\) the number of candidate bit-widths,
\(C\) the number of calibration batches, \(r\) the number of Hutchinson
probes, and \(Q\) the number of greedy bit-reduction steps. 
For our method, each calibration batch requires one forward-backward pass,
which provides gradients for all layers simultaneously. Hence, gradient
collection costs \(\mathcal{O}(C C_{\mathrm{FB}})\), where
\(C_{\mathrm{FB}}\) is the cost of one forward-backward pass. The resulting
gradients are reused across all \(K\) candidate bit-widths, giving a cost
of \(\mathcal{O}(KN)\). The greedy allocation examines \(L\) layers at each
of \(Q\) reduction steps, resulting in \(\mathcal{O}(QL)\). Thus,
\[
\mathcal{C}_{\mathrm{ours}}
=
\mathcal{O}(C C_{\mathrm{FB}}+KN+QL).
\]
In contrast, HAWQ-V2 estimates the Hessian trace separately for each layer
using \(r\) Hutchinson probes. If \(C_{\mathrm{HVP}}\) denotes the cost of
one Hessian-vector product, this requires
\(\mathcal{O}\!\left(L(C_{\mathrm{FB}}+rC_{\mathrm{HVP}})\right)\).
The subsequent candidate-bit evaluation and greedy allocation remain
\(\mathcal{O}(KN)\) and \(\mathcal{O}(QL)\), respectively. Hence,
\[
\mathcal{C}_{\mathrm{HAWQ}}
=
\mathcal{O}\!\left[
L(C_{\mathrm{FB}}+rC_{\mathrm{HVP}})+KN+QL
\right].
\]
Thus, our method avoids the per-layer second-order HVP computations and
instead uses a shared first-order gradient computation. This advantage is
particularly relevant for SNNs, where the computational graph is unrolled
over the simulation timesteps.

\section{Investigation on Fisher-Hessian Relationship in SNNs}

To understand the relationship between Fisher and Hessian based sensitivities, we perform a layerwise analysis using
\[
H_{ii} = \alpha_l F_{ii} + R_{i},
\]
Here, $H_{ii}$ and $F_{ii}$ are the diagonal Hessian and Fisher entries corresponding to parameter $i$, $\alpha_l$ is estimated independently for each layer and $R_i$ captures the remaining curvature. This decomposition is used \textbf{only for analysis} and is not required by our quantization procedure. 
\begin{figure}[!ht]
    \centering
    \includegraphics[width=\columnwidth]{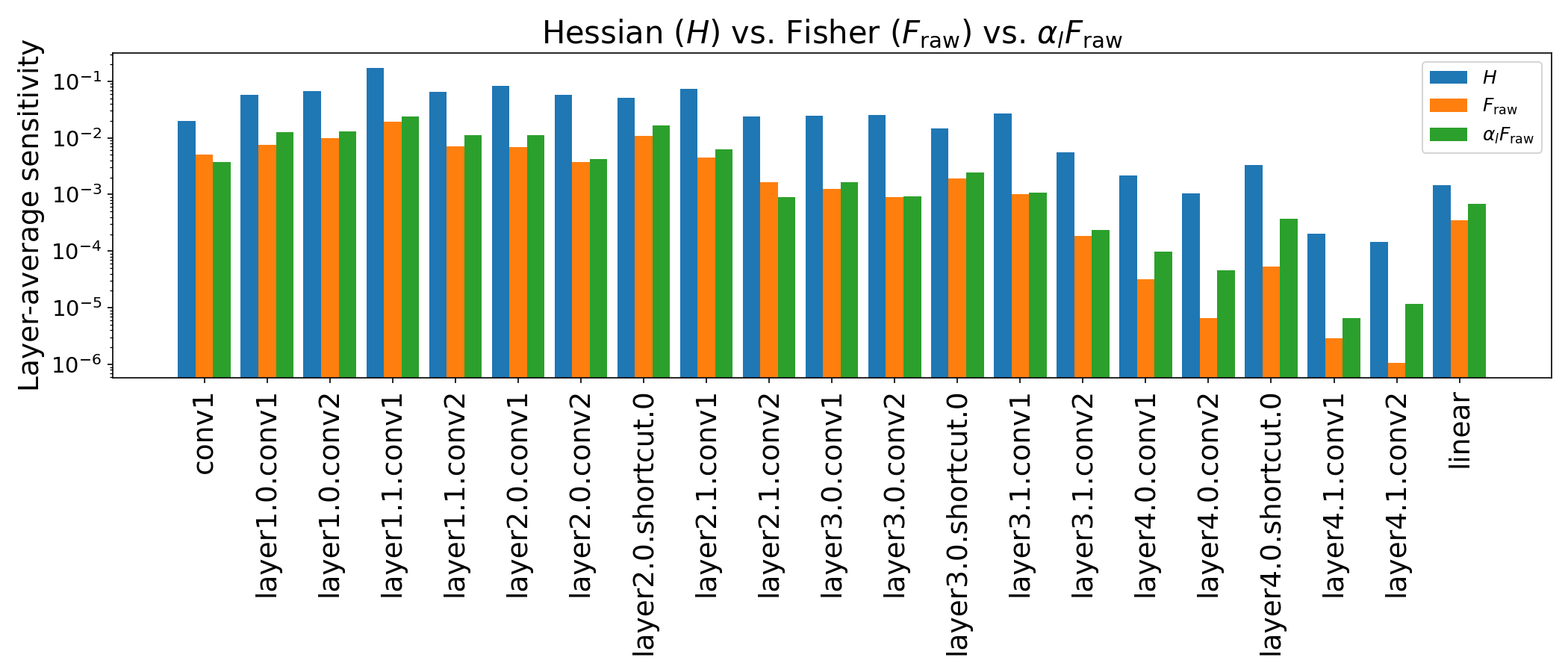}
    \caption{Raw values of Hessian, Fisher, and $\alpha_l . F$ for ResNet-18 on the CIFAR-10 dataset.}
    \label{fig:H_F}
\end{figure}
As depicted in Figure.~\ref{fig:H_F}, the raw layer-wise Hessian and Fisher sensitivities do not match; the Fisher values are generally much smaller than the corresponding Hessian values, and therefore the raw Fisher cannot be interpreted as a direct numerical approximation of the Hessian. Now, applying the layer-wise scaling $\alpha_l$ reduces this discrepancy, but $\alpha_l$ still does not fully match $H$, indicating that a simple scaling relationship alone is insufficient to explain the Hessian curvature.

\begin{figure}[!ht]
    \centering
    \includegraphics[width=\columnwidth]{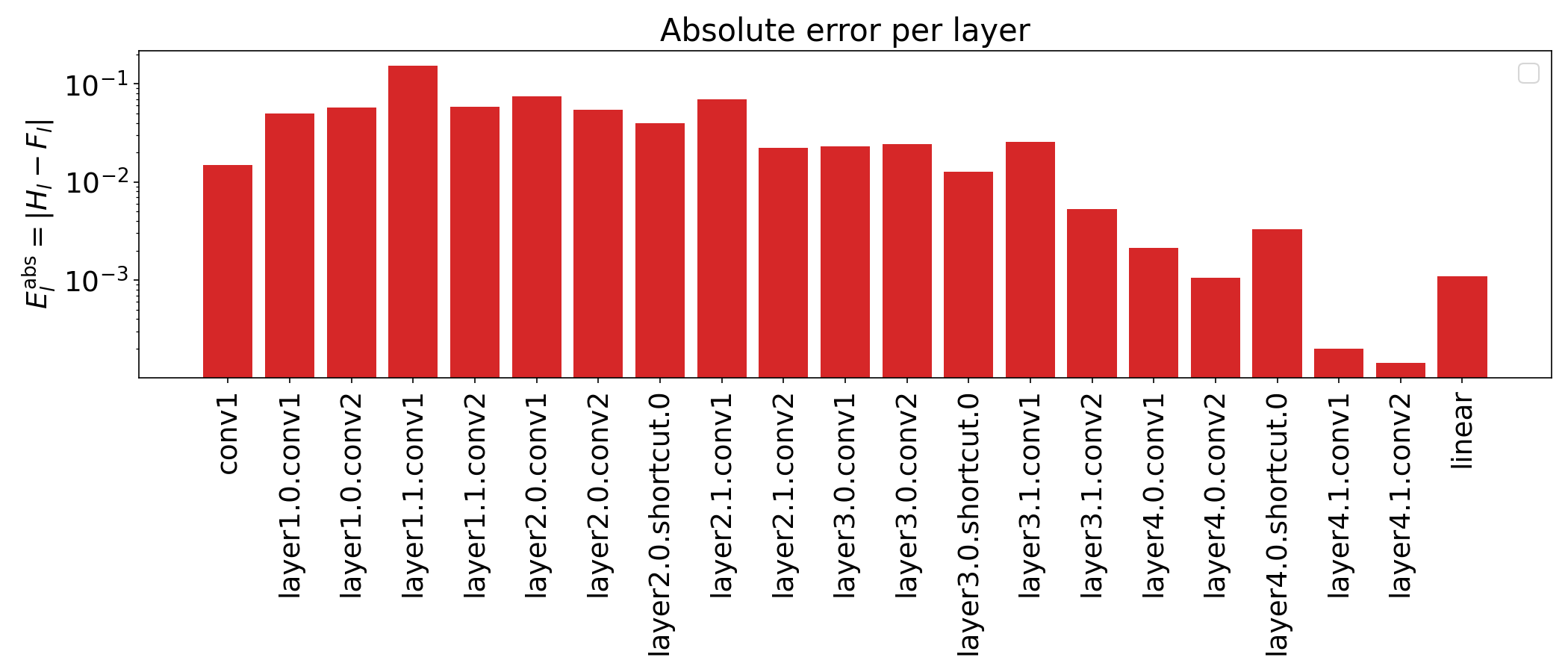}
    \caption{Layer-wise absolute error between the diagonal Hessian and raw Fisher sensitivities for ResNet-18 on CIFAR-10.}
    \label{fig:fisher_error_abs}
\end{figure}
\begin{figure}[!ht]
    \centering
    \includegraphics[width=\columnwidth]{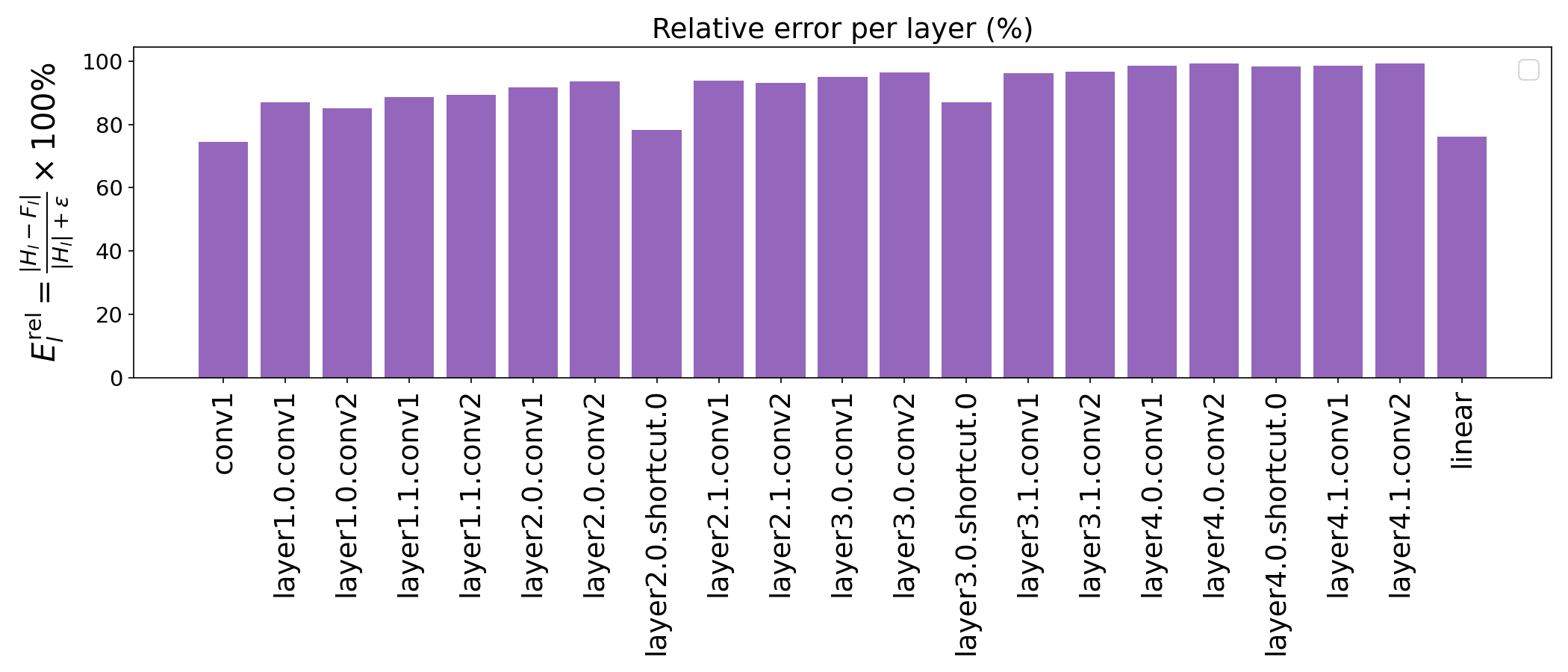}
    \caption{Layer-wise relative error between the diagonal Hessian and raw Fisher sensitivities for ResNet-18 on CIFAR-10.}
    \label{fig:fisher_error_rel}
\end{figure}

To quantitatively assess the discrepancy between Fisher and Hessian sensitivities, we compute the layer-wise absolute error
\[
E_l^{\mathrm{abs}} = |H_l-F_l|
\]
and relative error
\[
E_l^{\mathrm{rel}} =
\frac{|H_l-F_l|}{|H_l|+\epsilon}\times 100\%.
\]
As shown in Figure.~\ref{fig:fisher_error_abs} and Figure.~\ref{fig:fisher_error_rel}, the raw Fisher exhibits a substantial magnitude discrepancy with respect to the Hessian across several layers. In particular, the relative error indicates that the Fisher should not be interpreted as a numerically exact approximation of the diagonal Hessian under surrogate-gradient training.

Importantly, despite this difference in magnitude, the two criteria exhibit a strong agreement in their \textbf{relative layer-wise behavior}. The Pearson correlation between the Hessian and Fisher is $\textbf{0.90}$, while the Spearman rank correlation \cite{inbook} is $\textbf{0.89}$. In particular, the high Spearman correlation shows that layers identified as relatively more sensitive by the Hessian are also generally ranked highly by the Fisher, and vice versa. This is the property most relevant to our mixed-precision allocation, since the objective is to distinguish relatively sensitive layers rather than reproduce the exact numerical value of the Hessian.

\begin{figure}[!ht]
    \centering
    \includegraphics[width=\columnwidth]{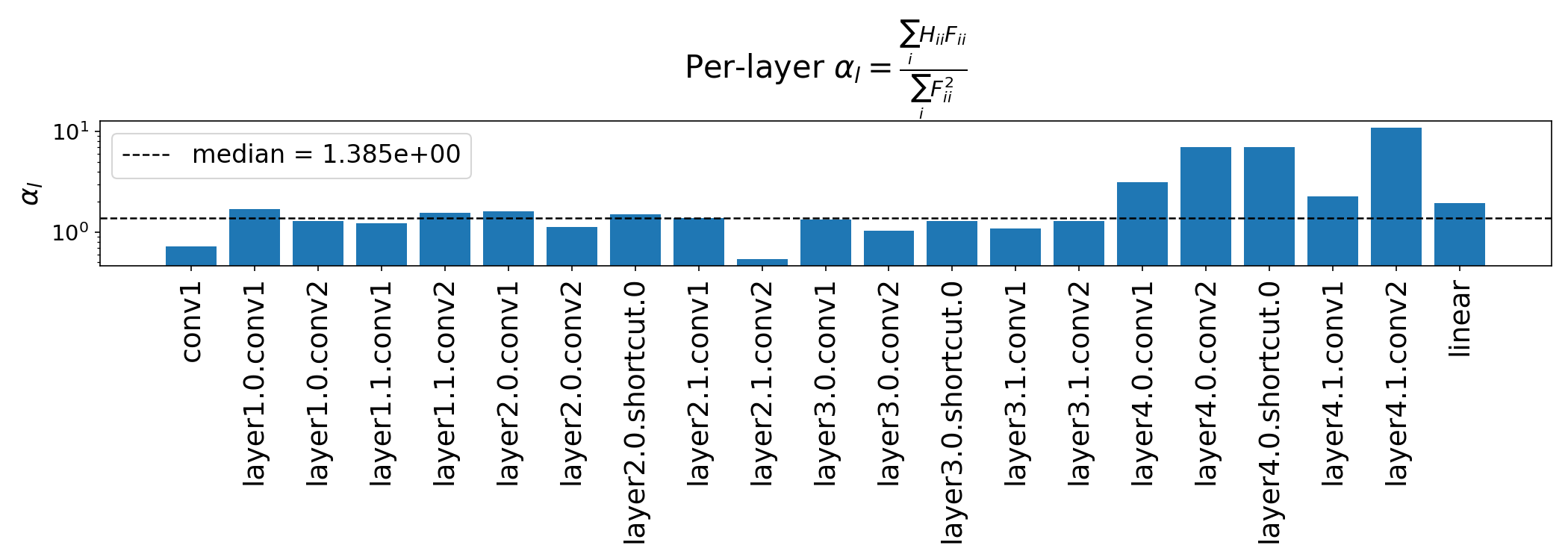}
    \caption{Values of $\alpha_l$ per layer for ResNet-18 on the CIFAR-10 dataset.}
    \label{fig:alpha_l}
\end{figure}

\textbf{Role of layer-wise scaling and residual curvature. }The variation of $\alpha_l$ over layers, shown in Figure.~\ref{fig:alpha_l}, further demonstrates that the Hessian–Fisher relationship is not governed by a single universal scale factor. The layer-dependent scaling varies considerably, confirming that the mismatch between $H$ and $F$ cannot be removed by simply applying one global multiplicative constant. Nevertheless, the strong rank agreement observed above indicates that this magnitude mismatch does not disrupt the relative sensitivity structure across layers. 
The remaining discrepancy is captured by the residual
\[
R_i = H_{ii} - \alpha.F_{ii}.
\]
\begin{figure}[!ht]
    \centering
    \includegraphics[width=\columnwidth]{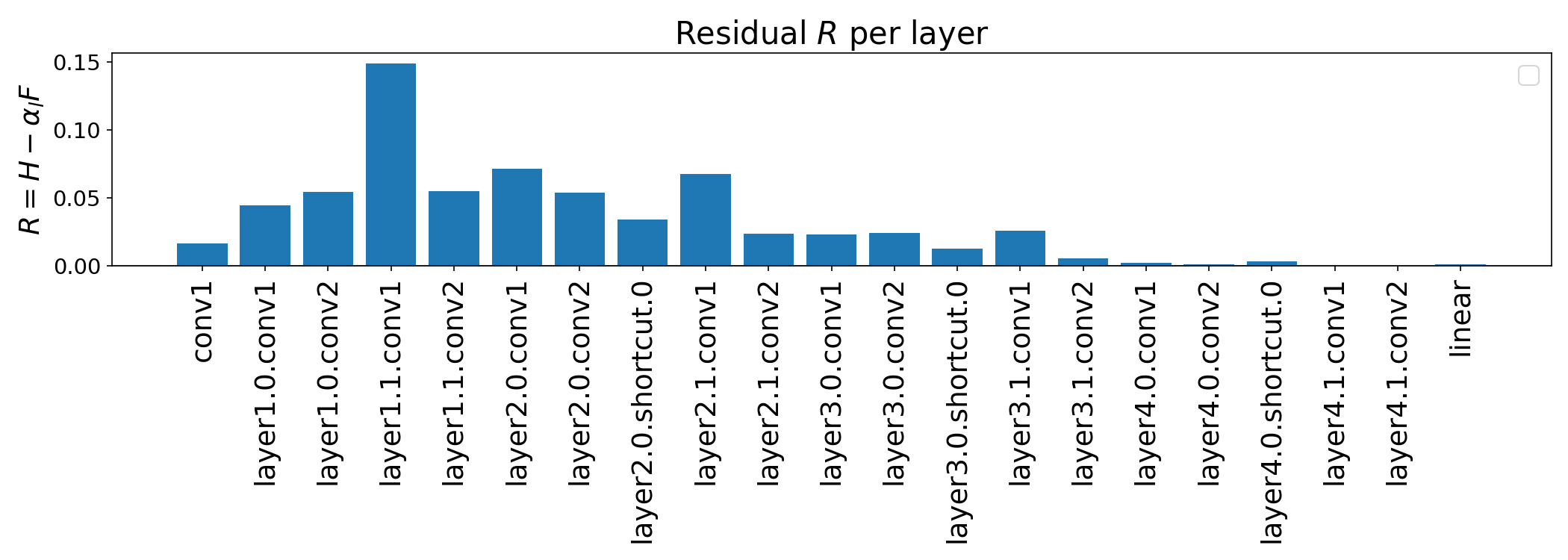}
    \caption{Values of $R_i$ per layer for ResNet-18 on the CIFAR-10 dataset.}
    \label{fig:residual}
\end{figure}
As shown in Figure.~\ref{fig:residual}, the residual is substantial for some layers and very small for others. This is also evident in Figure.~\ref{fig:H_F}, where $\alpha_lF$ remains separated from $H$ for several layers. However, after incorporating the residual, i.e., comparing $H$ with $\alpha_lF+R$, the reconstructed values closely match the Hessian across layers. This confirms that the observed discrepancy can be viewed as a combination of a Fisher-aligned component and a layer-dependent residual curvature component.

The important observation is therefore not that $F$ numerically equals $H$, but that \textbf{the Fisher preserves much of the relative sensitivity structure of the Hessian even when their absolute magnitudes differ substantially}. The high Pearson and Spearman correlations support this observation, while the $\alpha_l$ and residual analyses explain why the numerical values do not directly coincide. This provides an empirical justification for using Fisher sensitivity as a practical surrogate for Hessian sensitivity in our bit-allocation procedure, without requiring Hessian computation itself.

\section{Ablation Studies on Surrogate Gradients}
We further investigate whether the choice of surrogate gradient affects the proposed mixed-precision bit allocation strategy. Since surrogate gradients determine the backward-pass approximation of the non-differentiable spiking function, different surrogate functions can potentially alter the estimated layer-wise sensitivity and, consequently, the resulting bit-width assignment. To assess the robustness of our approach to this choice, we evaluate five commonly used surrogate gradients---Atan, Sigmoid, FastSigmoid, Piecewise Linear, and Triangular---on a ResNet-18~\cite{NEURIPS2021_afe43465} obtained through the QCFS conversion framework on the CIFAR-10 dataset. For each surrogate gradient, we independently compute the layer-wise sensitivity scores used by our bit allocation method and derive the corresponding mixed-precision configuration under the same overall bit-budget. This allows us to isolate the effect of the surrogate gradient while keeping the network architecture, conversion framework, dataset, and allocation procedure fixed. We then compare the resulting bit-width assignments and compression performance across the five surrogate-gradient choices to determine whether the proposed allocation strategy remains consistent under different backward-pass approximations.

% We further investigate whether the choice of surrogate gradient affects the proposed mixed-precision bit allocation strategy. Specifically, we evaluate five commonly used surrogate gradients - Atan, Sigmoid, FastSigmoid, Piecewise Linear, and Triangular on a ResNet-18~\cite{NEURIPS2021_afe43465} obtained through the QCFS conversion framework on the CIFAR10 dataset.
\begin{figure}[!ht]
    \centering
    \includegraphics[width=\columnwidth]{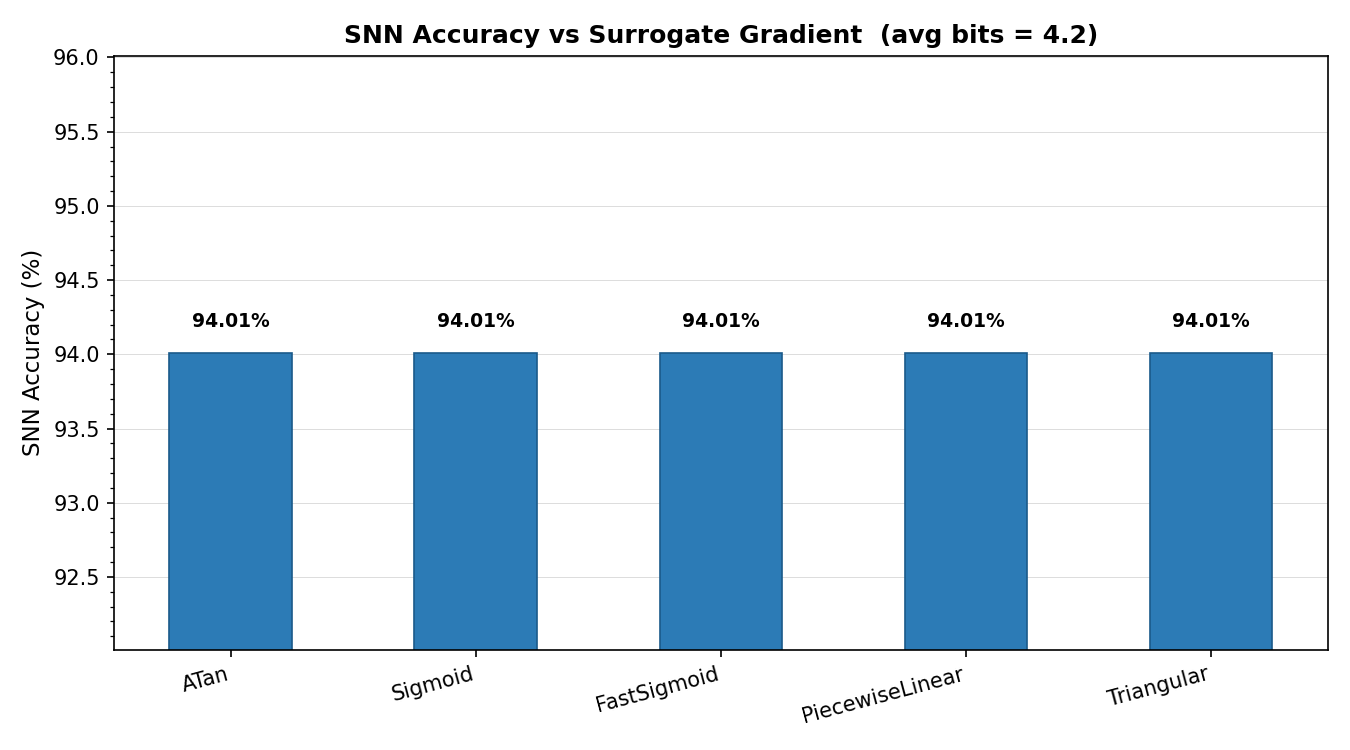}
    \caption{Quantized accuracy vs. surrogate gradient approximations. The model used here is ResNet-18 trained on CIFAR-10~\cite{li2017cifar10} using the QCFS~\cite{bu2022optimal} framework.}
    \label{fig:acc_surrogate1}
\end{figure}
\subsection{Coarse Bit-width Search Space (\texorpdfstring{$\{4,8\}$}{4,8})} 
We first evaluate the influence of different surrogate gradients using a coarse candidate bit-width set of \{4,8\}. Figure~\ref{fig:acc_surrogate1} shows that all five surrogate gradient produce identical quantized models, achieving the same accuracy of 94.01\%. This suggests that the proposed algorithm is largely insensitive to the choice of the surrogate gradient when only two candidate bit-width are available.
\begin{figure}[!ht]
    \centering    \includegraphics[width=0.9\columnwidth]{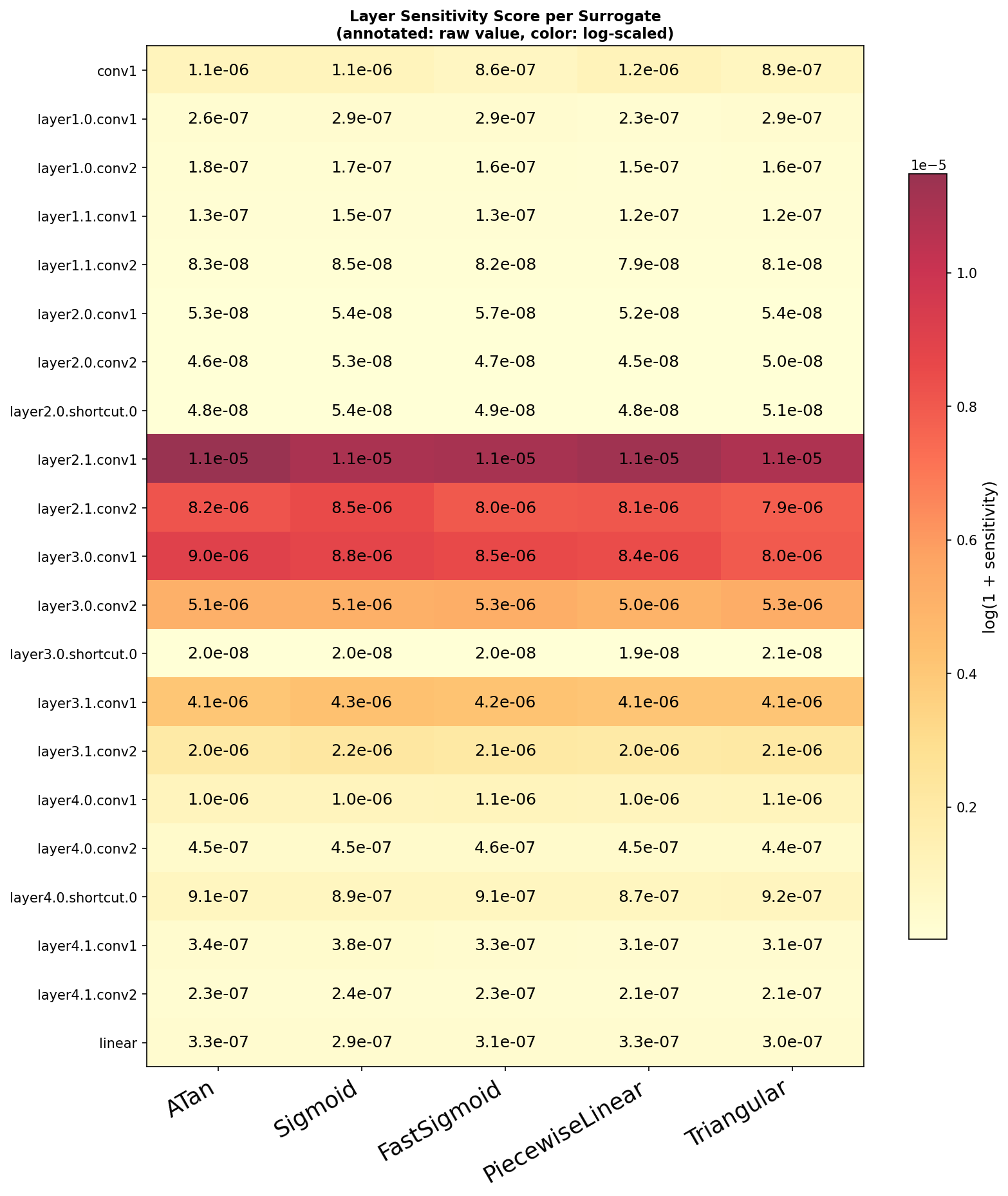}
    \caption{Sensitivity of each layer of ResNet-18 SNN trained on CIFAR-10~\cite{CIFAR} using the QCFS~\cite{bu2022optimal} framework.}
\label{fig:sensitvity_gradient_supp}
\end{figure}
We analyze the layer-wise sensitivity values computed using different surrogate gradients. As shown in Figure~\ref{fig:sensitvity_gradient_supp}, although the absolute sensitivity scores differ slightly across surrogate functions, the overall sensitivity pattern remains consistent. In particular, the same groups of layers are consistently identified as highly sensitive, while the remaining layers exhibit only minor numerical variations.
\begin{figure}[!ht]
    \centering
    \includegraphics[width=0.9\columnwidth]{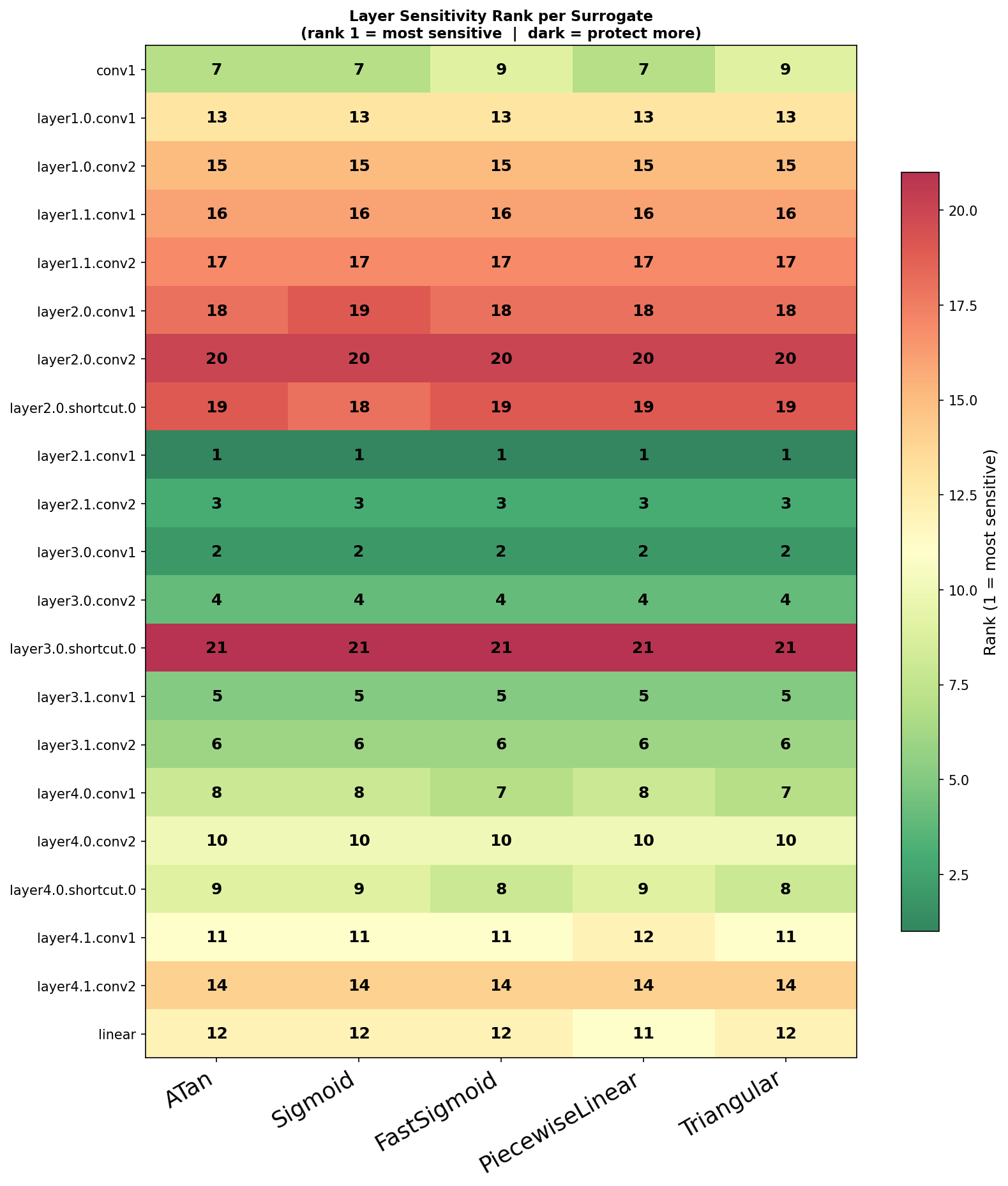}
    \caption{Comparison of layer rankings obtained using different surrogate gradients for ResNet-18~\cite{NEURIPS2021_afe43465} on CIFAR-10~\cite{CIFAR} using the QCFS~\cite{bu2022optimal} framework.}
    \label{fig:ranking_gradient1}
\end{figure}
\begin{figure}[!ht]
    \centering    \includegraphics[width=0.9\columnwidth]{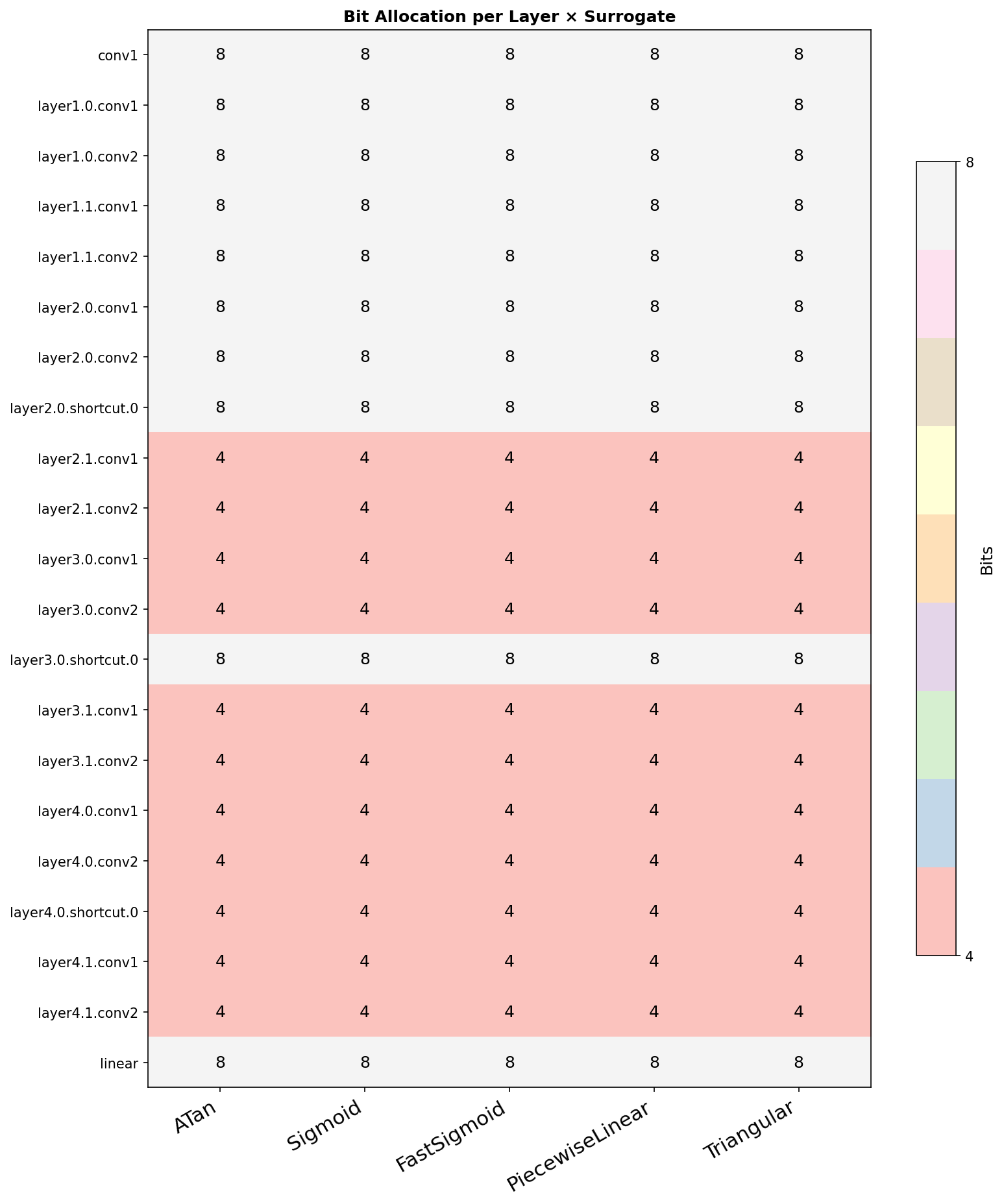}
    \caption{Layer-wise bit allocation of ResNet-18~\cite{NEURIPS2021_afe43465} SNN using different surrogate functions.}
    \label{fig:surrogate_bit_allocation1}
\end{figure}

Since the proposed algorithm depends primarily on the relative importance of layers rather than their absolute sensitivity values, we next compare the resulting layer rankings.
Figure~\ref{fig:ranking_gradient1} shows that the ranking produced by different surrogate gradients are highly consistent, with most layers retaining identical ranks and only a few layers exchanging neighboring positions. 
Figure~\ref{fig:surrogate_bit_allocation1} presents the bit allocation obtained using different surrogate gradient functions. Although each surrogate produces slightly different gradient magnitudes and corresponding sensitivity scores, the resulting layer rankings remain consistent. Because bit allocation is performed over a small set of candidate precisions, these small score differences do not affect the selected bit-width for any layer. Therefore, all surrogate gradients generate an identical mixed-precision configuration, which in turn results in identical model accuracy, as shown in Figure~\ref{fig:acc_surrogate1}.

\subsection{Fine-Grained Candidate Bit-width Setting (\texorpdfstring{$\{1,2,4,8\}$}{1,2,4,8})}

\begin{figure}[!ht]
    \centering
    \includegraphics[width=0.9\columnwidth]{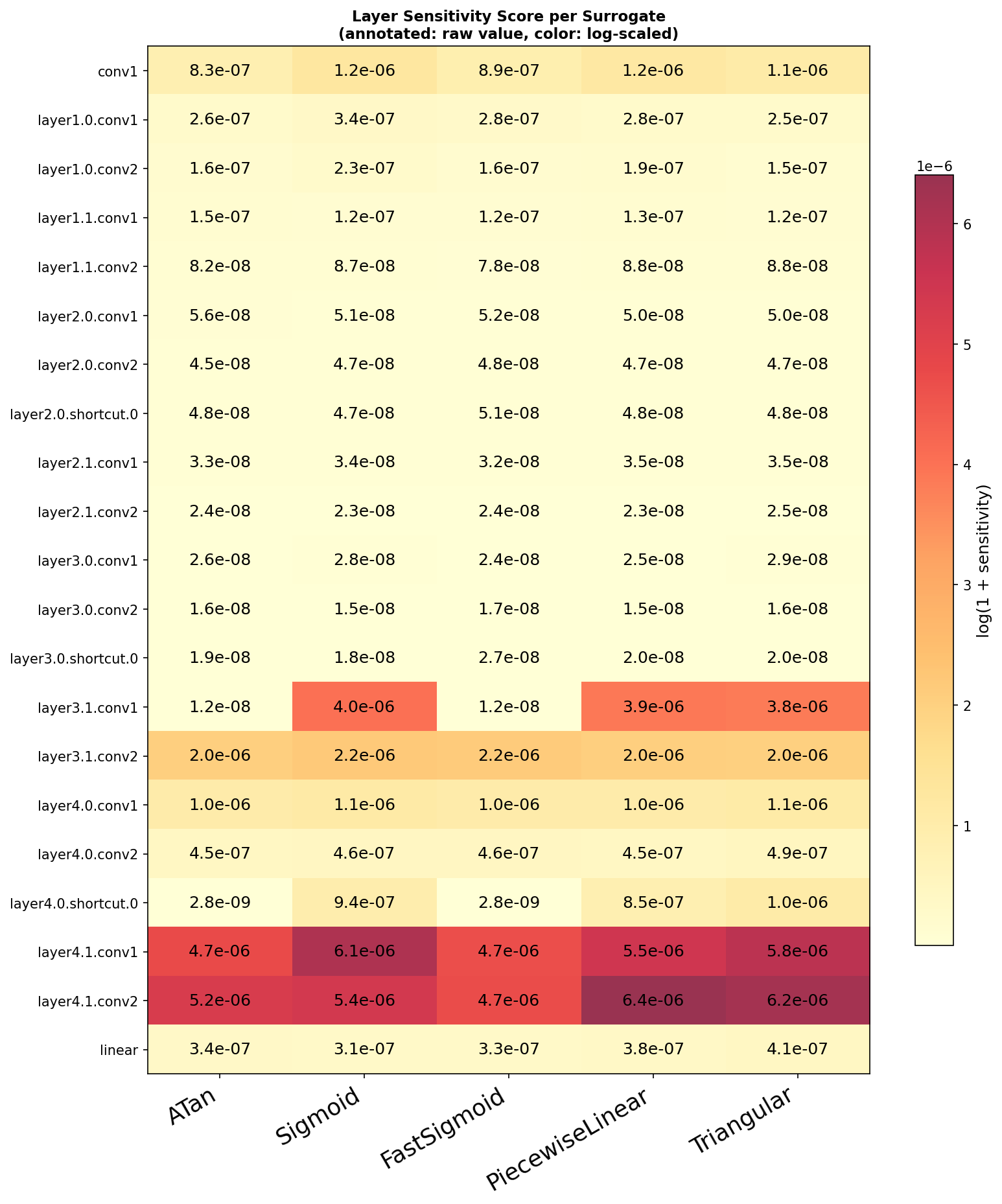}
    \caption{Layer-wise sensitivity of ResNet-18 SNN trained on CIFAR-10 using QCFS~\cite{bu2022optimal} framework. Candidate bits used here are $\{1,2,4,8\}$.}
    \label{fig:sensitivity_gradient2}
\end{figure}

Unlike the coarse candidate bit-width setting discussed in the previous subsection, the finer search space of $\{1,2,4,8\}$ gives the algorithm more flexibility in assigning layer-wise precisions. Consequently, variations in the surrogate functions influence the sensitivity ordering of layers, resulting in different mixed-precision configurations.  Figure~\ref{fig:sensitivity_gradient2} illustrates the layer sensitivity scores obtained using different surrogate gradients. 

Although all surrogate functions identify the final residual block as the most quantization-sensitive region of the network, noticeable differences are observed in the sensitivity values across layers. These differences suggest that surrogate gradients produce slightly different estimates of layer importance when a richer candidate precision set is available.
\begin{figure}[!ht]
    \centering
    \includegraphics[width=0.9\columnwidth]{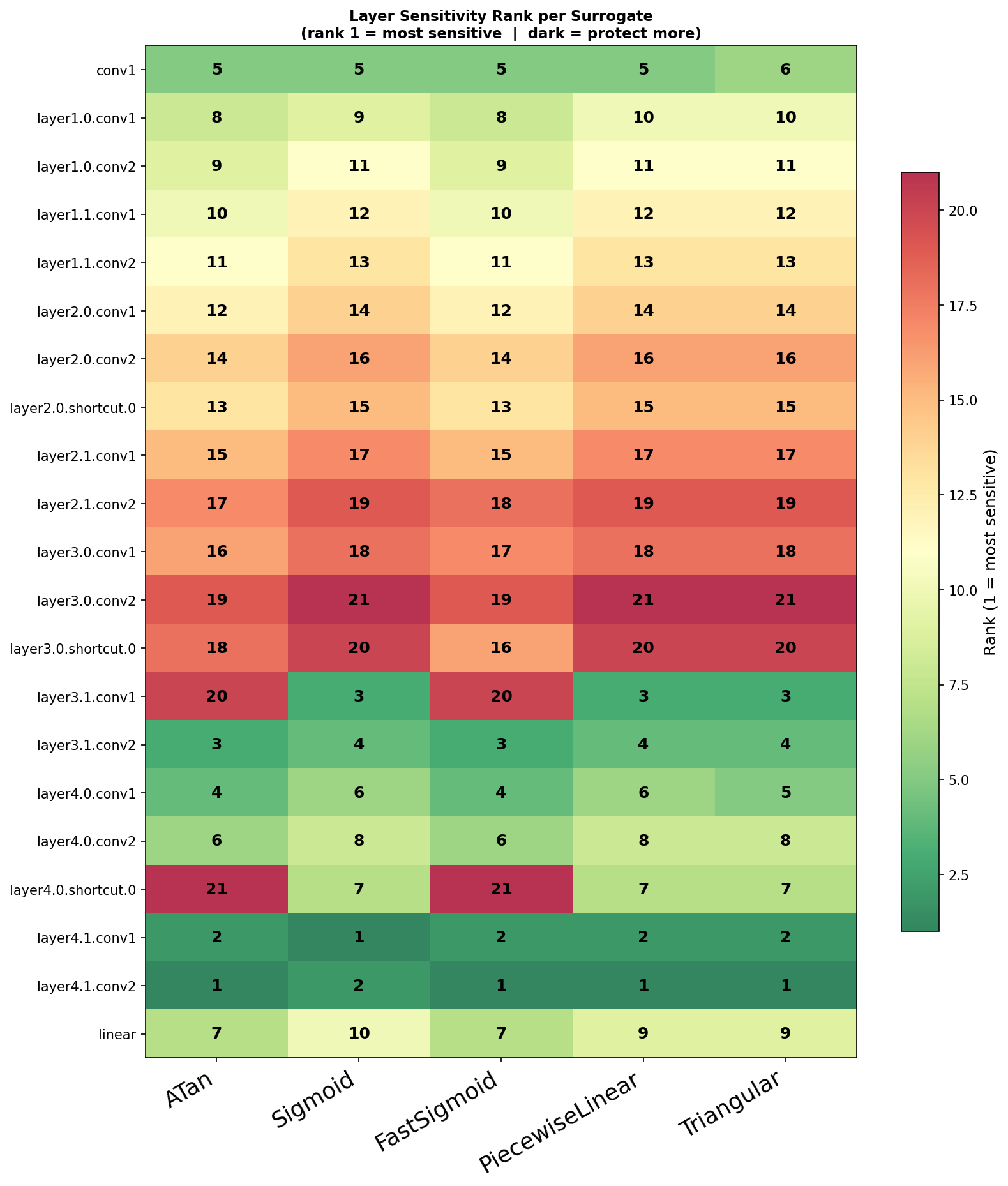}
    \caption{Comparison of layer sensitivity rankings obtained using different surrogate gradients for ResNet-18 SNN on CIFAR-10~\cite{CIFAR} using the QCFS~\cite{bu2022optimal} framework.}
    \label{fig:layer_ranking2}
\end{figure}
\begin{figure}[!ht]
    \centering
    \includegraphics[width=0.9\columnwidth]{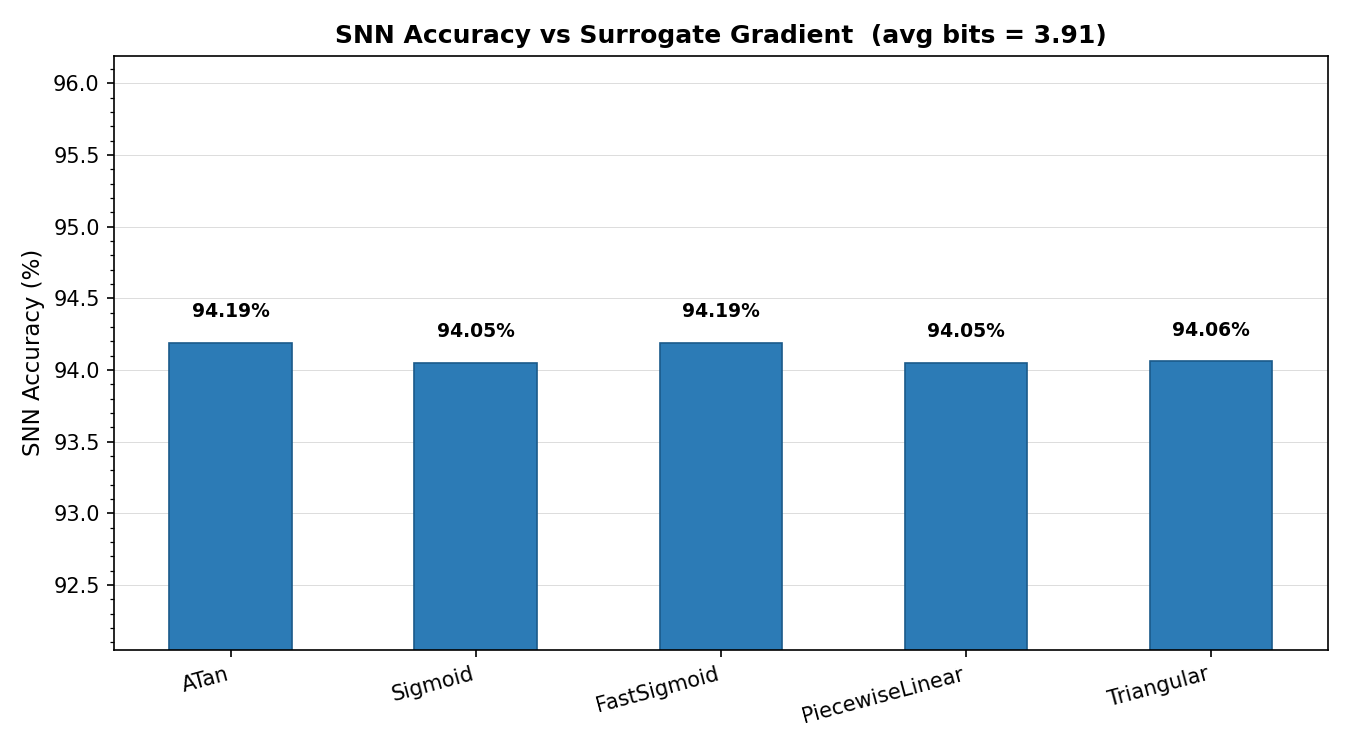}
    \caption{Quantized accuracy versus surrogate gradient functions for ResNet-18 SNN trained on CIFAR-10~\cite{CIFAR} dataset using the QCFS~\cite{bu2022optimal} framework.}
    \label{fig:acc_surrogate2}
\end{figure}
To better understand these variations, Figure~\ref{fig:layer_ranking2} converts the sensitivity values into layer-wise rankings. Compared to $\{4,8\}$ setting, the agreement among surrogate gradients decreases considerably. Several layers undergo substantial change in rank, especially in the later stages of the network where sensitivity values are relatively close. Unlike the coarse $\{4,8\}$ search space, the richer candidate bit-width set provides the optimization algorithm with greater flexibility to distinguish between layers with similar sensitivities, allowing these ranking differences to influence the final precision agreement.
% These ranking differences are reflected in the subsequent bit allocation, where layers whose relative sensitivity changes across surrogate gradients are assigned different precisions. In contrast, layers that consistently receive similar sensitivity scores retain the same bit-width across all surrogate functions. It is worth noting that only a small subset of layers exhibit changes in their assigned precision, while the majority of layers preserve identical bit-widths across all surrogate gradients. This indicates that the overall mixed-precision configuration remains highly consistent despite local variations in the sensitivity estimates. Consequently, the proposed bit allocation algorithm responds to surrogate-specific sensitivity differences while maintaining a stable allocation for layers whose importance is consistently identified.
These ranking differences are reflected in the bit allocation, with layers whose sensitivity changes across surrogate gradients receiving different precisions. In contrast, consistently ranked layers retain the same bit-width. Since only a small subset of layers change precision, the overall mixed-precision configuration remains highly consistent, indicating that the proposed algorithm adapts to surrogate-specific sensitivity while maintaining stable allocations for consistently important layers.

\begin{figure}[!ht]
    \centering
    \includegraphics[width=\columnwidth]{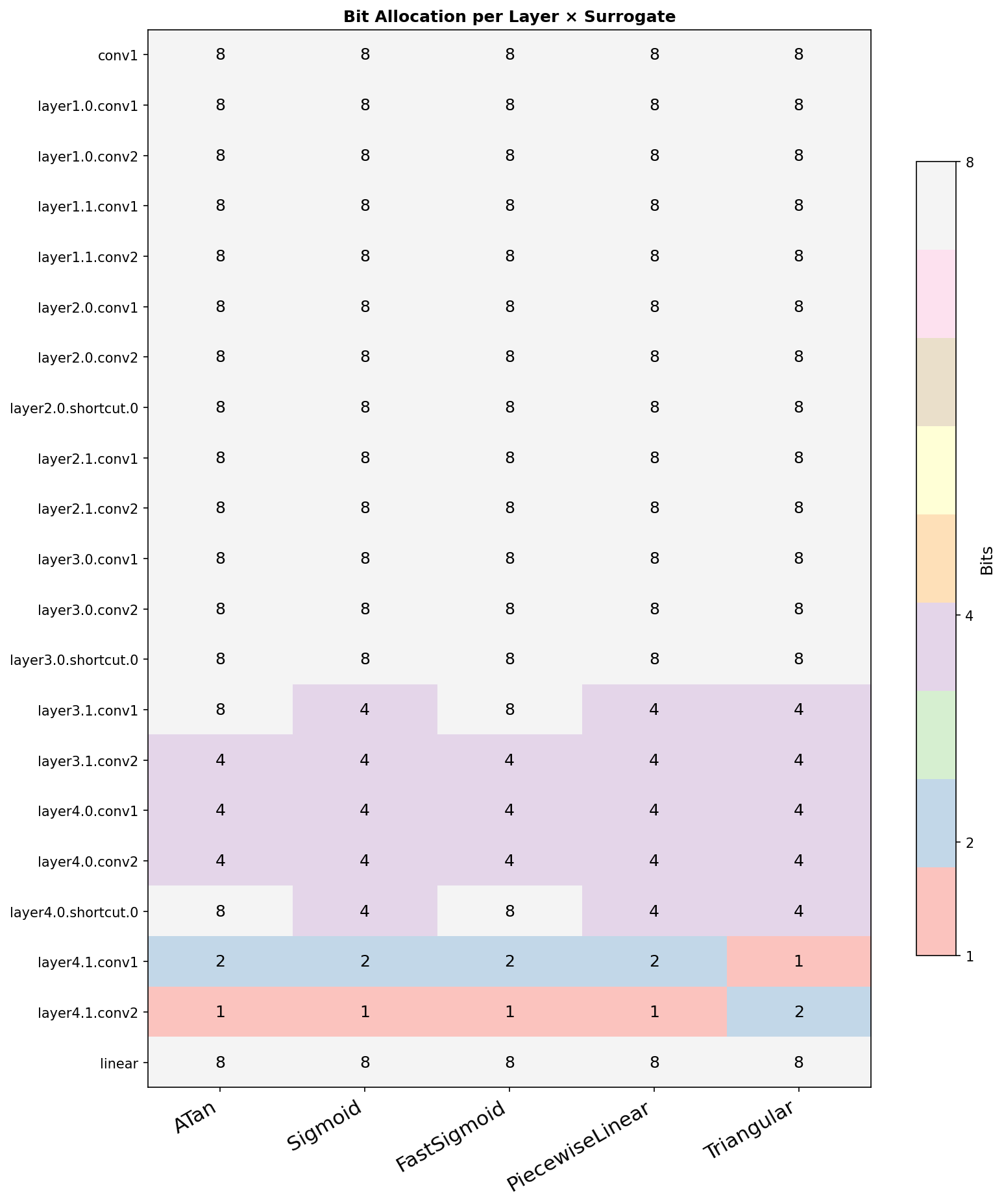}
    \caption{Layer-wise bit allocation of ResNet-18~\cite{NEURIPS2021_afe43465} SNN using different surrogate gradient functions.}
    \label{fig:bit_allocation2}
\end{figure}

The influence of surrogate-specific ranking differences on the final mixed-precision configuration is illustrated in Figure~\ref{fig:bit_allocation2}. Consistent with the observed ranking variations, the finer candidate bit-width set allows the algorithm to assign different precisions to a few layers across surrogate gradients, while the majority of layers retain identical bit-widths. This confirms that the proposed bit allocation strategy directly follows the surrogate-specific layer sensitivity estimates. 
The final quantized accuracy remains remarkably stable, as shown in Figure~\ref{fig:acc_surrogate2}. The accuracy varies only from 94.05\% to 94.19\% across all surrogate gradients, demonstrating that the proposed mixed-precision framework is robust to the choice of surrogate gradient even when a richer candidate bit-width search space is employed.

\section{Ablation studies on Protection Factor Parameter \texorpdfstring{$\beta$}{beta}}
The protection factor is defined as 
\[
P_l = 1 + \beta(1-m_l),
\]
where $\beta$ controls the contribution of threshold-margin information. We evaluate $\beta \in [0,2]$ on ResNet-18 and VGG-16 using CIFAR-10 and CIFAR-100, while keeping all other settings fixed.
\begin{figure}[ht]
    \centering
    \includegraphics[width=\columnwidth]{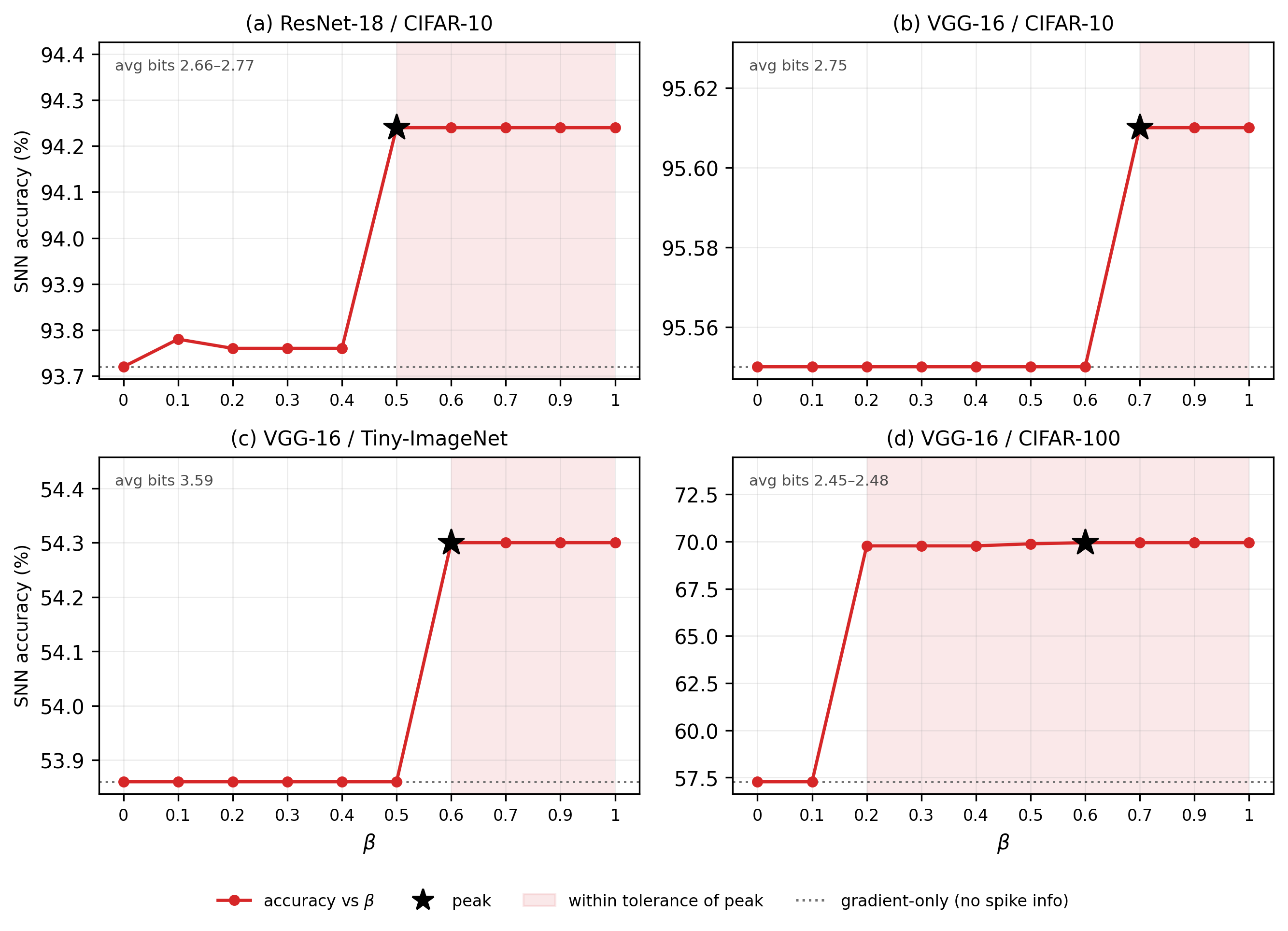}
    \caption{Beta analysis of the protection factor for different models trained on different datasets.}
    \label{fig:beta_analysis}
\end{figure}
As shown in Figure.\ref{fig:beta_analysis}, introducing the threshold-margin term $(\beta > 0)$ improves quantized accuracy by $57.5\%-71.78$ on VGG-16 (CIFAR-100) over the gradient-only setting $(\beta = 0)$. The optimal $\beta$ varies across architectures and datasets but a broad high-accuracy region is observed in all cases. Based on these results, we use $\beta=0.7$ as a fixed default value in our main experiments rather than tuning it seperately for each configuration.

\section{Layer-wise Bit Allocation Comparison}

To further analyze the proposed mixed-precision framework, we compare the layer-wise bit allocations generated by our gradient-based sensitivity metric with those obtained using HAWQ-V2. We report the assigned bit-width for every quantized layer across multiple SNN and ANN architectures. This comparison provides a qualitative assessment of whether the proposed first-order sensitivity metric captures similar layer importance as the second-order Hessian-based criterion.

\begin{table}[!ht]
\centering
\vspace{2pt}
\small
\renewcommand{\arraystretch}{1.1}
\setlength{\tabcolsep}{4pt}
\definecolor{lightgreen}{RGB}{220, 240, 220}
\rowcolors{2}{white}{lightgreen}
\begin{tabular}{lcc}
\toprule
\textbf{Layer} & \textbf{Ours} & \textbf{HAWQ-V2} \\
\midrule
conv-1  & 8 & 8 \\
conv-2  & 8 & 8 \\
conv-3  & 8 & 8 \\
conv-4  & 8 & 8 \\
conv-5  & 8 & 4 \\
conv-6  & 8 & 8 \\
conv-7  & 8 & 4 \\
conv-8  & 4 & 8 \\
conv-9  & 4 & 4 \\
conv-10 & 4 & 4 \\
conv-11 & 4 & 4 \\
conv-12 & 4 & 4 \\
conv-13 & 4 & 4 \\
FC-1    & 4 & 4 \\
FC-4    & 4 & 4 \\
FC-7    & 8 & 8 \\
\bottomrule
\end{tabular}
\caption{Layer-wise bit allocation for VGG-16 SNN trained on CIFAR-10~\cite{CIFAR} using the QCFS~\cite{bu2022optimal} framework.}
\label{tab:vgg16_c10_bits}
\end{table}
Although the proposed method relies only on first-order gradient information, the resulting bit allocations closely resemble those produced by HAWQ-V2 for most layers. Differences primarily appear in layers whose sensitivities are relatively similar, where multiple bit assignments can satisfy the target compression budget.
\begin{table}[t]
\centering
\vspace{2pt}
\small
\setlength{\tabcolsep}{4pt}
\definecolor{lightgreen}{RGB}{220, 240, 220}
\rowcolors{2}{white}{lightgreen}
\begin{tabular}{lcc}
\toprule
\textbf{Layer} & \textbf{Ours} & \textbf{HAWQ-V2} \\
\midrule
conv-1  & 8 & 8 \\
conv-2  & 8 & 8 \\
conv-3  & 8 & 8 \\
conv-4  & 8 & 8 \\
conv-5  & 8 & 8 \\
conv-6  & 4 & 8 \\
conv-7  & 8 & 8 \\
conv-8  & 4 & 8 \\
conv-9  & 4 & 4 \\
conv-10 & 8 & 4 \\
conv-11 & 4 & 4 \\
conv-12 & 4 & 4 \\
conv-13 & 4 & 4 \\
FC-1    & 4 & 4 \\
FC-4    & 4 & 4 \\
FC-7    & 8 & 8 \\
\bottomrule
\end{tabular}
\caption{Layer-wise bit allocation for VGG-16 SNN trained on CIFAR-100~\cite{CIFAR} using the QCFS~\cite{bu2022optimal} framework.}
\label{tab:vgg16_c100_bits}
\end{table}
Tables~\ref{tab:vgg16_c10_bits} and~\ref{tab:vgg16_c100_bits} compare the layer-wise bit allocation produced by the proposed method and HAWQ-V2 for VGG-16 SNN trained on CIFAR-10 and CIFAR-100 datasets, respectively. The overall allocation patterns are consistent, indicating that the first-order sensitivity metric captures layer importance comparable to the Hessian-based criterion of HAWQ-V2. Notably, the proposed method assigns lower bit-widths to several convolutional layers while maintaining higher precision for the most sensitive layers, resulting in a more aggressive compression strategy. Despite these differences, the overall mixed-precision configurations remain similar.

\begin{table}[!ht]
\centering
\vspace{4pt}
\small
\setlength{\tabcolsep}{4pt}
\definecolor{lightgreen}{RGB}{220, 240, 220}
\rowcolors{2}{white}{lightgreen}
\begin{tabular}{lcc}
\toprule
\textbf{Layer} & \textbf{Ours} & \textbf{HAWQ-V2} \\
\midrule
conv-1 & 8 & 8 \\
conv-2 & 8 & 8 \\
conv-3 & 8 & 8 \\
conv-4 & 4 & 4 \\
conv-5 & 4 & 4 \\
conv-6 & 4 & 4 \\
conv-7 & 4 & 4 \\
conv-8 & 4 & 4 \\
FC-1   & 8 & 8 \\
\bottomrule
\end{tabular}
\caption{Layer-wise bit allocation obtained for VGGSNN trained on CIFAR10-DVS~\cite{li2017cifar10} using the TET~\cite{deng2022tet} framework.}
\label{tab:vggsnn_dvs_bits}
\end{table}

\begin{table*}[t]
\centering
\small
\setlength{\tabcolsep}{6pt}
\definecolor{lightgreen}{RGB}{220, 240, 220}
\rowcolors{2}{white}{lightgreen}
\begin{tabular}{llc}
\toprule
\textbf{Stage} & \textbf{Layer(s)} & \textbf{Weight Bit} \\
\midrule
Stem &
Conv2d\_1a\_3x3.conv -- Conv2d\_3b\_1x1.conv &
8 \\
&
Conv2d\_4a\_3x3.conv &
4 \\
\midrule
Mixed\_5b &
branch1x1.conv -- branch3x3dbl\_2.conv &
6 \\
&
branch3x3dbl\_3.conv &
4 \\
&
branch\_pool.conv &
8 \\
\midrule
Mixed\_5c &
branch1x1.conv -- branch3x3dbl\_1.conv &
6 \\
&
branch3x3dbl\_2.conv -- branch3x3dbl\_3.conv &
4 \\
&
branch\_pool.conv &
6 \\
\midrule
Mixed\_5d &
branch1x1.conv -- branch5x5\_1.conv &
6 \\
&
branch5x5\_2.conv &
4 \\
&
branch3x3dbl\_1.conv -- branch3x3dbl\_2.conv &
6 \\
&
branch3x3dbl\_3.conv &
4 \\
&
branch\_pool.conv &
6 \\
\midrule
Mixed\_6a &
branch3x3.conv &
1 \\
&
branch3x3dbl\_1.conv &
6 \\
&
branch3x3dbl\_2.conv -- branch3x3dbl\_3.conv &
4 \\
\midrule
Mixed\_6b -- Mixed\_6e &
All convolution layers &
4 \\
\midrule
Mixed\_7a &
branch3x3\_1.conv &
4 \\
&
branch3x3\_2.conv &
1 \\
&
branch7x7x3\_1.conv -- branch7x7x3\_3.conv &
4 \\
&
branch7x7x3\_4.conv &
1 \\
\midrule
Mixed\_7b &
branch1x1.conv &
4 \\
&
branch3x3\_1.conv -- branch3x3dbl\_2.conv &
1 \\
&
branch3x3dbl\_3a.conv &
4 \\
&
branch3x3dbl\_3b.conv &
1 \\
&
branch\_pool.conv &
4 \\
\midrule
Mixed\_7c &
branch1x1.conv &
1 \\
&
branch3x3\_1.conv &
2 \\
&
branch3x3\_2a.conv -- branch3x3\_2b.conv &
1 \\
&
branch3x3dbl\_1.conv &
2 \\
&
branch3x3dbl\_2.conv -- branch\_pool.conv &
1 \\
\midrule
Head &
fc &
1 \\
\bottomrule
\end{tabular}
\caption{Layer-wise bit allocation produced by the proposed gradient-based mixed-precision quantization method for Inception-v3 ANN on ImageNet~\cite{imagenet}. Consecutive layers that are assigned the same bit-width are grouped for concise reporting.}
\label{tab:inception_gradient_bits}
\end{table*}

% \bibliographystyle{splncs04}
% \bibliography{supplementary}

\end{document}